\documentclass[10pt,twocolumn,letterpaper]{article}

\usepackage[pagenumbers]{cvpr} 

\usepackage[dvipsnames,table]{xcolor}

\newcommand{\methodname}{DISEF\xspace}

\newcommand{\methodnamelongbold}{\textbf{D}iversified \textbf{I}n-domain \textbf{S}ynthesis with \textbf{E}fficient \textbf{F}ine-tuning\xspace}
\newcommand{\sample}{x}
\newcommand{\samplelabel}{y}
\newcommand{\classlabel}{c}
\newcommand{\numbershots}{K}
\newcommand{\numbersynthshots}{K_{syn}}
\newcommand{\numberclasses}{N}
\newcommand{\supportset}{\mathcal{X}}
\newcommand{\syntheticset}{\mathcal{X}'}
\newcommand{\syntheticsample}{x'}
\newcommand{\jointset}{\hat{\mathcal{X}}}
\newcommand{\model}{\mathcal{F}}
\newcommand{\visualfeatures}{f_v}
\newcommand{\textualfeatures}{f_t}
\newcommand{\textualfeaturesw}[1]{f_{t_#1}}
\newcommand{\captioner}{\mathcal{ICM}}
\newcommand{\generator}{\mathcal{G}}
\newcommand{\captionset}{C}
\newcommand{\captionsample}{p}
\newcommand{\latentvec}{z}
\newcommand{\zeroshot}{W_{zs}}

\newcommand{\lorainput}{i}
\newcommand{\loraoutput}{o}
\newcommand{\loraweight}{W}
\newcommand{\lorarank}{r}
\newcommand{\textmodel}{T}
\newcommand{\visionmodel}{V}
\newcommand{\SA}{\mathcal{SA}}
\newcommand{\Query}{\mathcal{Q}}
\newcommand{\Value}{\mathcal{V}}

\DeclareMathOperator*{\argmax}{arg\,max}

\definecolor{cvprblue}{rgb}{0.21,0.49,0.74}
\usepackage[pagebackref,breaklinks,colorlinks,citecolor=cvprblue]{hyperref}
\usepackage{multirow}
\usepackage{subcaption}
\usepackage{amsmath}
\usepackage{pifont} 
\usepackage{makecell}

\definecolor{azure(colorwheel)}{rgb}{0.0, 0.5, 1.0}

\usepackage{graphicx}
\usepackage[export]{adjustbox}
\newcommand{\cmark}{\textcolor{ForestGreen}{\ding{51}}}
\newcommand{\xmark}{\textcolor{red}{\ding{55}}}

\newcommand*\samethanks[1][\value{footnote}]{\footnotemark[#1]}

\def\paperID{5574} 
\def\confName{CVPR}
\def\confYear{2024}

\title{Diversified in-domain synthesis with efficient fine-tuning for few-shot classification}

\author{Victor G. Turrisi da Costa\thanks{\scriptsize{Victor G. Turrisi da Costa and Nicola Dall'Asen contributed equally.}} \,$^{\textcolor{azure(colorwheel)}{1}}$ \quad Nicola Dall'Asen\samethanks \, $^{\textcolor{azure(colorwheel)}{1,2}}$ \quad    Yiming Wang$^{\textcolor{azure(colorwheel)}{3}}$\\
Nicu Sebe$^{\textcolor{azure(colorwheel)}{1}}$ \quad Elisa Ricci$^{\textcolor{azure(colorwheel)}{1,3}}$ \\
\normalsize{$^{\textcolor{azure(colorwheel)}{1}}$ University of Trento \quad $^{\textcolor{azure(colorwheel)}{2}}$ University of Pisa\quad $^{\textcolor{azure(colorwheel)}{3}}$ Fondazione Bruno Kessler}
}

\begin{document}
\maketitle

\begin{abstract}
Few-shot image classification aims to learn an image classifier using only a small set of labeled examples per class. A recent research direction for improving few-shot classifiers involves augmenting the labelled samples with synthetic images created by state-of-the-art text-to-image generation models. Following this trend, we propose \methodnamelongbold (\textbf{\methodname}), a novel approach which addresses the generalization challenge in few-shot learning using synthetic data. \methodname consists of two main components. First, we propose a novel text-to-image augmentation pipeline that, by leveraging the real samples and their rich semantics coming from an advanced captioning model, promotes in-domain sample diversity for better generalization. Second, we emphasize the importance of effective model fine-tuning in few-shot recognition, proposing to use Low-Rank Adaptation (LoRA) for joint adaptation of the text and image encoders in a Vision Language Model. We validate our method in ten different benchmarks, consistently outperforming baselines and establishing a new state-of-the-art for few-shot classification.
Code is available at \url{https://github.com/vturrisi/disef}.
\vspace{-10pt}
\end{abstract}

\section{Introduction}\label{sec:intro}

Few-shot classification aims to develop models that can categorize new samples, \emph{i.e.} the \emph{query} set, into a set of classes by only learning from a very limited number of labeled samples of each class, \emph{i.e.} the \emph{support} set. This is especially relevant in application domains where collecting extensive labeled datasets is expensive or unfeasible. 
The main challenge in few-shot classification lies in how to learn generalizable representation from such a limited support set.
To address this issue, over the years, researchers have proposed different approaches, \textit{e.g.} based on meta-learning \cite{hospedales2020metalearning}, transfer learning \cite{zhuang2020comprehensive}, metric learning~\cite{jung2022few,li2023deep} and, more recently, on fine-tuning vision and language models (VLM) \cite{zhou2022conditional,zhou2023training,gao2021clipadapter,Shi_2023_ICCV}, complemented by data augmentation techniques.

\begin{figure}[!t]
    \centering
    \includegraphics[width=0.88\linewidth]{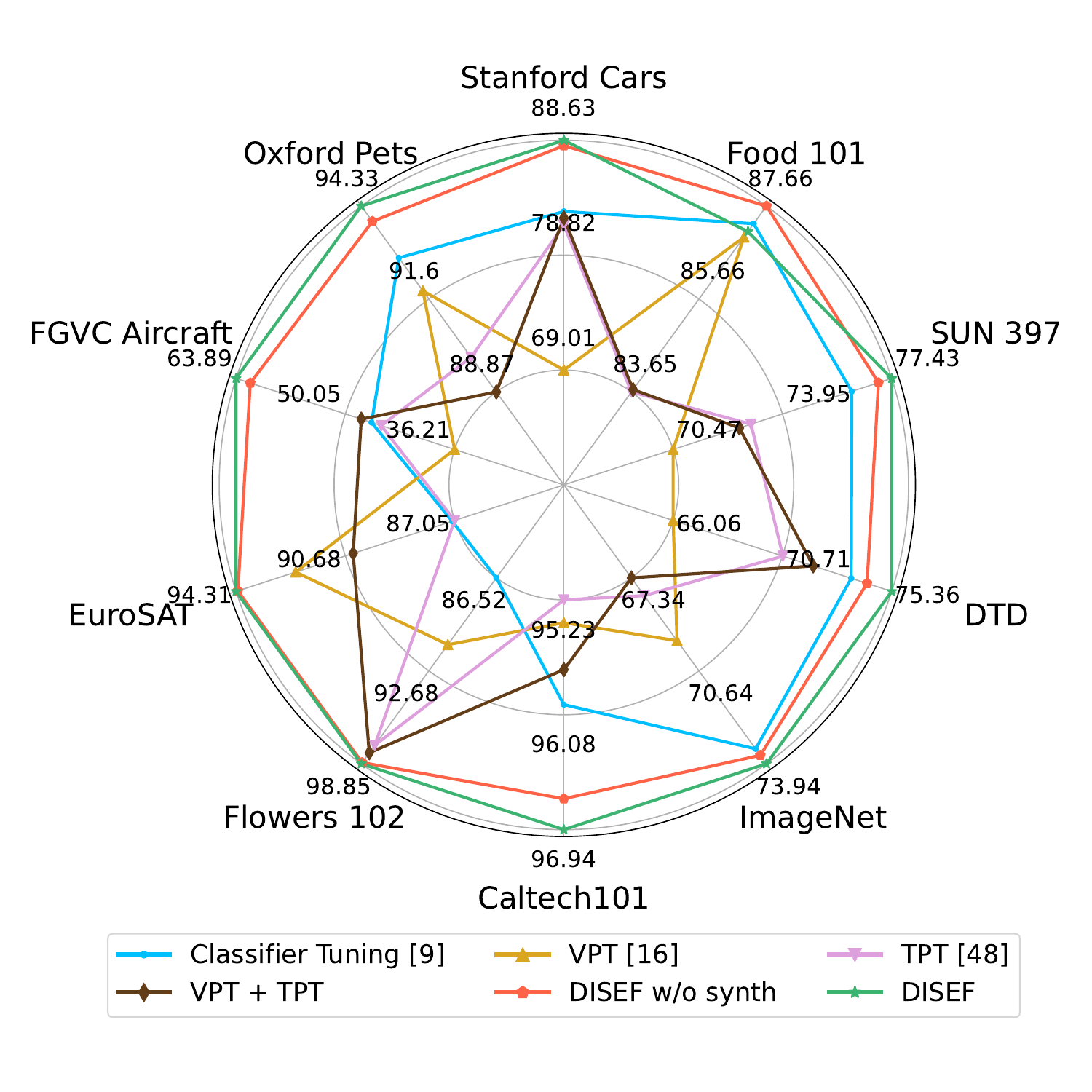}
    \vspace{-10pt}
    \caption{Radar chart comparing our proposed method, \methodname (with and without synthetic data) against other fine-tuning methods with Vision Language Models, namely Classifier Tuning \cite{he2023synthetic}, Text Prompt Tuning (TPT) \cite{Zhou_2022}, Visual Prompt Tuning (VPT) \cite{jia2022visual} and a combination of VPT and TPT. Different angles correspond to the ten different benchmark datasets.}
    \label{fig:teaser}
    \vspace{-10pt}
\end{figure}

Generative models and, in particular, text-to-image diffusion models \cite{sohl2015deep,ho2020denoising,rombach2022highresolution} have reached a significant level of maturity that enables the synthesis of highly photo-realistic images. These advances have spurred a new research trend that investigates the use of synthetic data in image recognition tasks \cite{tian2023stablerep,sariyildiz2023fake,zhou2023training}.
However, existing work mostly focuses on investigating novel generative models \cite{zhou2023training} to enrich the data with access to the distribution of the whole dataset, or on providing recipes to improve model pre-training \cite{tian2023stablerep}.

Surprisingly, how to use synthetic data to enhance classification models in data-scarce scenarios, \ie the few-shot learning setting, has not been well studied. He \etal~\cite{he2023synthetic} explored the idea of adopting synthetically generated data to increase the diversity of the support set, in order to learn classification models which are more robust and able to generalize better.
Lin \etal~\cite{lin2023explore} explored the potential of generated images to improve object detection in a few-shot scenario.
However, when adopting text-to-image generation models as in~\cite{he2023synthetic,lin2023explore, shipard2023diversity}, the process for creating synthetic data consists of simply requesting \texttt{a photo of a [CLASS]} without constraints on details.
This may yield out-of-domain synthetic images, \eg images of the correct class but with a very distinct viewpoint or visual style. Such diverse, but out-of-domain generation can degrade the classification performance~\cite{shipard2023diversity}.

Another aspect that so far has been overlooked in the literature of learning from limited real (and synthetic) data is related to strategies for fine-tuning pre-trained models. These strategies are fundamental to maximally boost the generalization capability of classifiers, especially in a few-show setting. 
The emergence of large vision and language models (VLM), such as CLIP~\cite{radford2021learning} has opened up different possibilities for performing model fine-tuning on downstream tasks. Notable approaches include text prompt tuning \cite{Zhou_2022, zhou2022conditional, Shi_2023_ICCV}, visual prompt tuning \cite{jia2022visual} or multi-modal prompt tuning \cite{Khattak_2023_CVPR}.
Recently, parameter-efficient fine-tuning (PEFT) methods \cite{rebuffi2017learning, pmlr-v97-houlsby19a, hu2021lora, li-liang-2021-prefix, zhang2023llamaadapter, gao2023llamaadapterv2, liu2022fewshot} have attracted attention, becoming the \textit{de facto} strategy for fine-tuning models in the text domain.
These approaches mainly consist of adding (smaller) adapters \cite{rebuffi2017learning, pmlr-v97-houlsby19a}, performing prompt tuning \cite{li-liang-2021-prefix, zhang2023llamaadapter, gao2023llamaadapterv2} or learning low-rank update matrices to the parameters \cite{hu2021lora}.
Nevertheless, these techniques, apart from prompt tuning, have seen limited adoption in computer vision, with only a few exceptions \cite{jia2022visual, chavan2023oneforall}.

In this paper, we propose to tackle the problem of few-shot classification with synthetic data by innovating key recipes in data synthesis and parameter-efficient model fine-tuning. Our approach, \methodnamelongbold (\textbf{\methodname}), brings two main contributions. 
First, we propose a \textit{novel text-to-image augmentation pipeline} which encourages in-domain data synthesis and promotes sample diversity.
Precisely, we leverage state-of-the-art captioning models, such as LLaVA~\cite{liu2023visual}, to produce textual descriptions of support images that are rich in semantic details. Such descriptions are then exploited as anchors in a cross-sample manner to promote in-domain synthesis with diversity.
We also incorporate real samples of the support set in the noise injection procedure of the diffusion-based generative model, so that the generated images have a consistent visual appearance to the real images of the same class. 
Second, we demonstrate that \textit{effective model fine-tuning} is a key factor in few-shot recognition and we, for the first time, leverage Low-Rank Adaptation (LoRA) \cite{hu2021lora} for jointly adapting the text and vision encoders of a VLM in the few-shot scenario.
Our approach provides the flexibility to choose which components of the VLM are updated without modifying the original architecture or the input of the networks, as we do not rely on learnable prompts in any of the modalities.
This represents a powerful yet simple adaptation strategy when learning in data-constrained scenarios.
We validate our proposed method on ten benchmarks for few-shot classification (see also Fig.~\ref{fig:teaser}), consistently outperforming baseline methods and setting a new state-of-the-art.

\noindent \textbf{The Contributions} of our work can be summarized as below:
\begin{itemize}
\item We introduce \methodname, a new framework for few-shot classification that leverages synthetic data and parameter-efficient fine-tuning.
\item For generating synthetic images, we propose a novel augmentation pipeline that leverages both support images and their captions for producing diverse but in-domain training samples.
\item For fine-tuning, we shed new light on the importance of model fine-tuning in the context of few-shot classification with VLMs and propose to leverage LoRA \cite{hu2021lora} for adapting both the vision and text encoders.
\item We achieve the new state-of-the-art for few-shot image classification on extensive benchmarks, proving the effectiveness of our proposed method.
\end{itemize}

\section{Related work}
We review related works on few-shot classification based on VLMs, the use of generative models for image data augmentation, and parameter-efficient fine-tuning methods.

\noindent\textbf{Few-shot recognition with VLMs.} Several recent works have shown the effectiveness of VLMs when applied to few-shot classification. For instance, Zhou \etal~\cite{Zhou_2022} proposed an adaptation method for VLMs that consists of adding learnable prompts to the text encoder, similar to a simple Text Prompt Tuning (TPT).
In \cite{zhou2022conditional}, this approach was further extended to condition the learnable text prompts on the input image.
Gao \etal~\cite{gao2021clipadapter} introduced CLIP-Adapter, which adds learnable linear layers to the output of both the vision and language encoders, freezing CLIP's parameters. Differently, Katthak \etal~\cite{Khattak_2023_CVPR} employed learnable prompts at multiple layers of the text and vision encoders, conditioning the visual prompts by linearly projecting the text prompts. Shi \etal~\cite{Shi_2023_ICCV} proposed LoGoPrompt, a method derived from the observations that images with written class names and natural images activate the same neurons in CLIP.
However, none of these works focused on enhancing the performance of few-shot classifiers derived from VLMs by using synthetic data.

\noindent\textbf{Synthetic data as additional training data.} 
In the last few years, diffusion models~\cite{sohl2015deep,ho2020denoising} have emerged as a powerful approach for generating highly realistic images. By driving the generation process with text~\cite{rombach2022highresolution} or with any other conditioning signals~\cite{zhang2023adding} (depth maps, pose, semantic maps, etc.) diffusion models have also demonstrated high flexibility.
These recent advances in image generation have encouraged researchers to investigate the use of synthetic images within recognition tasks, with the purpose of enriching the original data distribution. 
For instance, Tian \etal~\cite{tian2023stablerep} showed that it is possible to pre-train large models in a self-supervised way by using only synthetic data generated by Stable Diffusion~\cite{rombach2022highresolution}, achieving competitive performance with models pre-trained with real data.
Zhou \etal~\cite{zhou2023training} introduced 
Diffusion Inversion, an approach that uses a pre-trained Stable Diffusion model to create synthetic datasets. It ensures coverage of the original data manifold while producing novel samples that complete the training domain by generating variations of real samples, thus facilitating generalization.
Azizi \etal~\cite{azizi2023synthetic} found that fine-tuning Imagen~\cite{saharia2022photorealistic} on ImageNet leads to synthetic data that better matches the training data distribution. They also showed that by jointly using real and synthetic data the accuracy of a model on ImageNet can be improved. 
Shipard \etal~\cite{shipard2023diversity} showed how a recognition model trained only on synthetic data can generalize well on real data.
More recently, He \etal~\cite{he2023synthetic} proposed to leverage GLIDE~\cite{pmlr-v162-nichol22a} to generate synthetic data in a few- and zero-shot scenario. They also implemented a filtering scheme to eliminate samples from one class that exhibit close proximity to another class in the feature space. 
This work is the most closely related to ours, as they specifically used VLMs and a synthetic data generation approach for improving few-shot classification performance. 
However, their language enhancement pipeline, a fine-tuned T5 model~\cite{roberts2019exploring}, does not provide semantically rich captions, but only \textit{grammatically correct} sentences. Differently from them, we start from captions obtained by an image captioning model, thus providing the text-to-image pipeline with a grounded and detail-rich description. 
Additionally, the use of GLIDE, a two-stage pipeline, as the generative backbone makes the model impractical for larger datasets, especially given the fact they require up to 800 images per class.
These reasons, united with the controllable guidance coming from real images, make our generation pipeline more suitable for the few-shot scenario. 

\noindent\textbf{Parameter efficient fine-tuning methods.}
The advent of large deep learning models and their widespread use in several applications has prompted numerous endeavors to develop techniques for fine-tuning these models on downstream tasks. For instance, in
\cite{rebuffi2017learning} and \cite{pmlr-v97-houlsby19a} small trainable MLPs are added within the layers of a frozen pre-trained model. LLaMA-Adapter~\cite{zhang2023llamaadapter} and LLaMA-Adapter v2~\cite{gao2023llamaadapterv2} used learnable prompts at different layers of a transformer architecture, with the effect of progressively injecting information into the main model using a zero-initialized attention mechanism. Jia \etal \cite{jia2022visual} introduced Visual Prompt Tuning (VPT), a method inspired by the prompt tuning literature from the text domain. It adds learnable visual prompts in either a shallow way, where the prompts are added only at the input of the model, or a deep way, where multiple different prompts are employed in each layer. Hu \etal \cite{hu2021lora} introduced LoRA, a method which learns low-rank update matrices to adapt a pre-trained model. Chavan \etal \cite{chavan2023oneforall} proposed GLoRA, a modification of LoRA which uses a different set of learnable parameters per layer and leverages an evolutionary algorithm to select which learnable parameters to add per layer. 
In this work, we explore for the first time the use of LoRA to concurrently adapt the text and vision encoders of a VLM in the context of few-shot recognition.
Unlike previous approaches, we show that we do not need to adapt only a single modality, nor do we need to restrict the adaptation to simple learnable prompts.
This offers additional flexibility in terms of what can be used to adapt VLMs in the few-shot scenario, hopefully opening an avenue to connect fine-tuning methods for large language models (LLM) to fine-tuning methods for VLMs in data-constrained scenarios.

\section{Proposed approach}
\label{sec:approach}

\begin{figure*}[!t]
    \centering
    \includegraphics[width=0.80\linewidth]{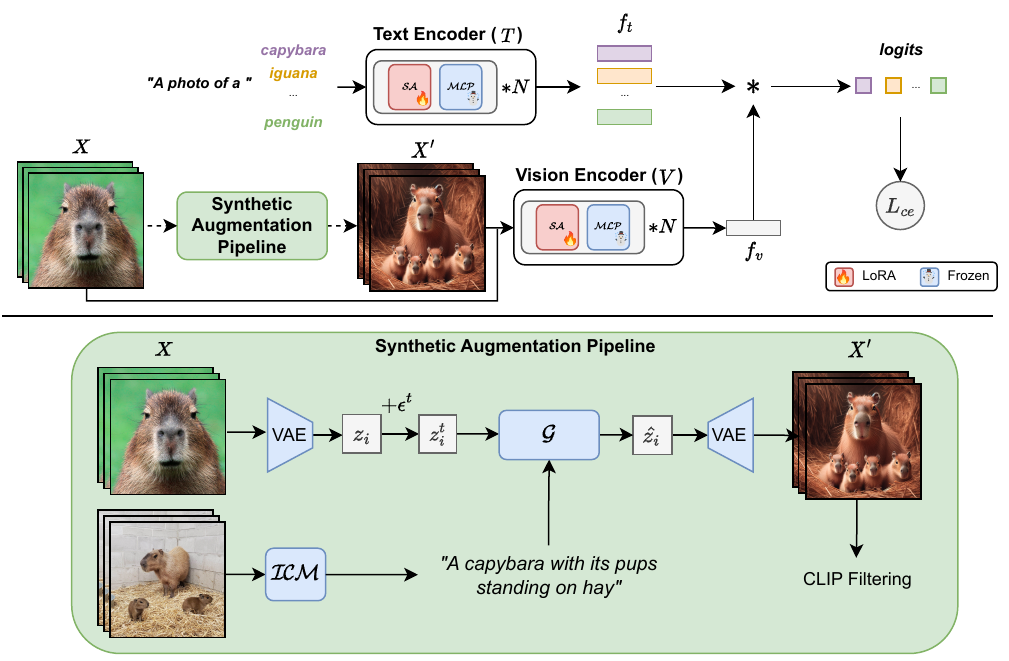}
    \vspace{-5pt}
    \caption{Proposed method for few-shot learning. At the top, we present our adaptation strategy. Starting from one of the few-shot images available, we generate additional training data by applying our Synthetic Augmentation Pipeline (SAP). Then, we treat both the real images and the synthetic images in the same manner. Considering an image $\sample$, we forward it through our vision encoder ($V$) to produce visual features $f_v$. In parallel, we forward all the class labels, combined with a pre-defined prompt template through the text encoder ($T$), generating text features $f_t$. Then, we compute the $logits = sim(f_v, f_t)$, where $sim$ is the cosine similarity function, and the cross-entropy loss $L_{ce}$. Finally, instead of updating all the parameters of our model, we modify the original model by adding LoRA layers in the query ($\mathcal{Q}$) and value ($\mathcal{V}$) embeddings of the self-attention ($\mathcal{SA}$) layers in both the text and vision encoders
    At the bottom, we show the SAP procedure. Starting from the set of images $\supportset$, we caption them with an image captioning model ($\mathcal{ICM}$), while also projecting them in the Stable Diffusion latent space. We inject noise into the latent vectors and run the reverse diffusion process with shuffled, per class, captions, obtaining synthetic images $\syntheticset$. Lastly, we filter $\syntheticset$ with CLIP to retain only synthetic images which are classified as their intended class.}
    \label{fig:method}
    \vspace{-10pt}
\end{figure*}

We consider the support set $\supportset = \{(\sample, \samplelabel)\}^{\numbershots, \numberclasses}$, containing tuples in the format of an image $\sample$ and its label $\samplelabel$, where $\numbershots$ is the number of images per class (\ie the number of shots), and $\numberclasses$ the number of classes. 
The goal is to learn a function ${\model}_\theta(\sample) \rightarrow \samplelabel$ that maps the image $\sample$ to its corresponding label $\samplelabel$.
Usually, $\model_\theta(\cdot)$ is represented by a neural network parametrized by $\theta$. 
Moreover, methods addressing few-shot image classification often exploit a pre-trained model and restrain the training to just a small subset of parameters \cite{Zhou_2022,zhou2022conditional,Khattak_2023_CVPR,Shi_2023_ICCV}.
Following a similar philosophy, we build our method \methodname~(as shown in Figure \ref{fig:method}) on top of a pre-trained VLM, where we fine-tune only a small fraction of the model with a novel application of parameter-efficient fine-tuning and a new Synthetic Augmentation Pipeline (SAP).

SAP is built on top of Stable Diffusion~\cite{rombach2022highresolution} with the objective of promoting diversified in-domain sample generation.
Different from~\cite{he2023synthetic}, which only uses one image and an augmented prompt at a time to generate synthetic data, SAP generates a set of synthetic images $\syntheticset$ by leveraging the whole support set together with their captions, which are rich in semantic details.
Specifically, given the support subset $\supportset_{\samplelabel}$ corresponding to a class label $\samplelabel$, we extract its image captions with an off-the-shelf image captioning model $\captioner$, obtaining a set of captions $\captionset_{\samplelabel}$. 
To encourage in-domain generation, we start the generation by embedding each real sample $\sample_i$ in the latent space of Stable Diffusion. 
Then, we inject noise to perturb the low-level details while maintaining high-level class semantics. Meanwhile, to encourage diversity, we condition the generation with the caption of another sample $\sample_j \in \supportset_{\samplelabel}$. In this way, we can generate diverse samples that are semantically correct and visually in-domain. 
Furthermore, to reduce the number of incorrect images, we filter them based on their similarity to the textual representation of their desired class.

At adaptation time, we treat both synthetic and real samples in a similar way, and we use the joint set $\jointset = \supportset \cup \syntheticset$ to fine-tune the model. Specifically, we add LoRA \cite{hu2021lora} layers to the query ($\Query$) and value ($\Value$) embeddings on the self-attention ($\SA$) layers of both vision and the text encoders.
This allows us to adapt a pre-trained VLM for both modalities, while at the same time being efficient.

\subsection{Synthetic Augmentation Pipeline}
SAP augments the support set $\supportset$ with additional synthetic data points that are interpolated within the domain inferred by $\supportset$. 
The in-domain synthesis requires the synthesized images not only to belong to the same semantic class but also to exhibit similar visual patterns as the real images. 
On the other hand, diversity is needed for the data to be useful for training and requires the synthesized images to present different semantic details.
Our proposed SAP encourages such in-domain diversity from two novel perspectives: first, by manipulating the visual representation of real images during the diffusion process, and second, by leveraging semantically detailed captions provided by a captioning model. 

\noindent\textbf{Condition on real images.} For a real image $\sample_i \in \supportset_{\samplelabel}$, we leverage a pre-trained VAE~\cite{rombach2022highresolution} to project the image into the latent space, obtaining a latent vector $\latentvec_i$.
Then, Gaussian noise $\epsilon^t$ is added to perturb the low-level visual details without changing the semantic class, obtaining a noisy latent vector $\latentvec_i^t$, for a given step $t$, following the diffusion process. Note that the amount of noise injected can be dataset-dependant as it correlates to the granularity among classes, \textit{i.e.} the more coarse-grained the classes are, the more noise we can inject without modifying the semantic class.

\noindent\textbf{Condition on semantic-detailed captions.} Given the support set $\supportset_{\samplelabel}$ of class $\samplelabel$, we first generate their captions using the image captioning model $\captioner$ obtaining a set of captions $\captionset_{\samplelabel}$. These add more details such as size, color, composition, and action, that are far richer than the standard CLIP prompt of \texttt{a photo of a [CLASS]}.

\noindent\textbf{Synthetic image generation.} We input $\latentvec_i^t$ to the generator $\generator$, together with a randomly sampled caption $\captionsample_j \in {\captionset}_{\samplelabel}, j \neq i$. 
The generation follows the classifier-free guidance procedure of diffusion models~\cite{ho2022classifier}, where we obtain a reconstructed latent vector $\hat{\latentvec_i} = \generator(\latentvec_i^t, \captionsample_j)$. We then decode $\hat{\latentvec_i}$ to obtain the synthetic image. By using $\captionsample_j$ and the noisy latent vector $\latentvec_i^t$ to condition the generation, we are more likely to generate visually in-domain samples corresponding to the same class, with different, yet plausible, semantic details. 

\noindent\textbf{Synthetic image filtering.} 
We further apply a filtering strategy to ensure the generated samples are aligned to the semantic class.
More formally, we leverage the zero-shot classifier $\zeroshot \in \mathbb{R}^{\numberclasses\times d}$ from CLIP for a particular dataset, where $d$ is the CLIP dimensionality.
First, we compute the latent representation for the generated images by forwarding them through CLIP's vision encoder, obtaining $\visualfeatures = \visionmodel(\syntheticsample)$ for each $\syntheticsample \in \syntheticset$. 
We obtain the predicted class for each sample $\hat{y} = \argmax (\visualfeatures \, \zeroshot^\intercal)$, where $\visualfeatures$ and $\zeroshot$ are $L_2$ normalized.
Lastly, we remove from $\syntheticset$ the samples whose $\hat{y}$ is not the correct class.

At each generation, we choose a random real image in the support set $\supportset$ and a random caption corresponding to a different image of the same class, followed by the sample filtering procedure. We repeat the generation for each class until we obtain a fixed number of synthetic samples $\numbersynthshots$ for each class, forming the final synthetic set $\syntheticset$.

\subsection{Parameter-efficient VLM fine-tuning}
\label{subsec:adaptation}
Inspired by recent advances in LLM fine-tuning, we propose to integrate LoRA~\cite{hu2021lora}, originally proposed for the text domain, to fine-tune a VLM for few-shot learning. 
Specifically, LoRA proposes to add small, low-rank update matrices to adapt a pre-trained model. Given a pre-trained dense layer $\loraoutput = \loraweight \,\lorainput$, where $\loraweight \in \mathbb{R}^{m\times n}$ is the weight matrix, $\lorainput$ is the input and $\loraoutput$ is the output, LoRA modifies the layer by adding an update matrix $\Delta\loraweight$, that can be further decomposed, as:
\begin{equation}
\loraoutput = \loraweight\,\lorainput + \Delta\loraweight\,\lorainput = \loraweight\,\lorainput + B\,A\,\lorainput,
\end{equation}
where $B \in \mathbb{R}^{m \times \lorarank}$, $A \in \mathbb{R}^{\lorarank \times n}$ are two low-rank matrices, $m$ and $n$ are the number of rows and columns in the original weight matrix $\loraweight$, and $\lorarank$ is the rank of the update matrix, with $\lorarank \ll \min(m, n)$. 
This decomposition greatly reduces the number of trainable parameters to just a fraction of the original layer with no impact on inference time, as it is possible to merge $\loraweight$ and $B\,A$ during inference.
$A$ is initialized by sampling from a Gaussian distribution and $B$ is initialized with zeros to avoid disrupting the model so that it produces the same results as the original model at the beginning of the fine-tuning process.
Lastly, $\Delta\loraweight\,\lorainput$ is further scaled with a weight $\frac{\alpha}{\lorarank}$, where $\alpha$ is a hyperparameter.

As the domain shift between the pre-training data and the few-shot task is not specific to a single modality, we propose to add LoRA layers to both the vision encoder $\visionmodel$ and the text encoder $\textmodel$, instead of only the text encoder. 
Specifically, we integrate the LoRA layers to the query ($\Query$) and value ($\Value$) embeddings of the self-attention ($\SA$) layers, the most effective placement as demonstrated in~\cite{hu2021lora}.

\subsection{Training}
Considering a single image $\sample$, we first forward it through the vision encoder $\visionmodel$ producing visual features $\visualfeatures = \visionmodel(\sample)$.
In parallel, we forward each class label $\classlabel$ with a default prompt through the text encoder $\textmodel$, producing $\textualfeatures = \{\textmodel(\classlabel)\}^N$. 
This default prompt is selected on a per-dataset basis, borrowing the same prompts in CLIP~\cite{radford2021learning}.
Then, we compute the logits as:
\begin{equation}
    logits = sim(\visualfeatures, \textualfeatures),
\end{equation}
where $sim(\cdot)$ is the cosine similarity function, $\textualfeatures = \{\textualfeaturesw{1}, \textualfeaturesw{2}, ..., \textualfeaturesw{\numberclasses}\}$ and $\numberclasses$ is the total number of classes.

Lastly, a cross-entropy loss $L_{ce}$ is used to compute the gradients of the model.
As our batch is composed of both real and synthetic images, we compute a separate loss for each part, $L_{real}$ and $L_{syn}$, respectively. 
We compute the final loss as the weighted average of the two cross-entropy losses $L_{real}$ and $L_{syn}$:
\begin{equation}
L_{ce} = \lambda L_{real} + (1 - \lambda) L_{syn}, \text{ where } \lambda \ge 0.5
\label{eq:loss}
\end{equation}

\section{Experiments}
\label{sec:experimental_setup}

\begin{table*}[!t]
    \centering
    \caption{16-shot results on all the datasets averaged across 3 seeds. We highlight \colorbox{ForestGreen!35}{\textbf{best}} and \colorbox{YellowOrange!35}{\underline{second best}} results.}
    \vspace{-8pt}
    \resizebox{1.0\textwidth}{!}{
    \begin{tabular}{lcccccccccc|c}
        \hline
        & Caltech101 & DTD & EuroSAT & FGVC Aircraft & ImageNet & Oxford Pets & Stanford Cars & SUN397 & Food 101 & Flowers 102 & Avg \\
        \hline
        
        Classifier Tuning~\cite{he2023synthetic} & 96.01 & 73.64 & 87.13 & 46.73 & 73.41 & 92.81 & 82.55 & 76.16 & \cellcolor{YellowOrange!35}\underline{87.28} & 86.52 & 80.22 \\

        Visual Prompt Tuning~\cite{jia2022visual} & 95.40 & 66.06 & 92.33 & 36.21 & 69.57 & 91.84 & 69.01 & 70.47 & 86.99 & 90.95 & 66.31 \\
        
        Text Prompt Tuning~\cite{Zhou_2022} & 95.23 & 70.73 & 87.05 & 45.50 & 67.97 & 89.89 & 81.40 & 72.95 & 83.65 & 97.62 & 79.20 \\
        
        VPT + TPT & 95.75 & 72.02 & 90.42 & 48.03 & 67.34 & 88.87 & 81.99 & 72.58 & 83.70 & 98.12 & 79.88 \\ \midrule
        
        \methodname w/o synth & \cellcolor{YellowOrange!35}\underline{96.71} & \cellcolor{YellowOrange!35}\underline{74.31} & \cellcolor{YellowOrange!35}\underline{94.25} & \cellcolor{YellowOrange!35}\underline{62.09} & \cellcolor{YellowOrange!35}\underline{73.64} & \cellcolor{YellowOrange!35}\underline{93.88} & \cellcolor{YellowOrange!35}\underline{88.17} & \cellcolor{YellowOrange!35}\underline{77.01} & \cellcolor{ForestGreen!35}\textbf{87.66} & \cellcolor{YellowOrange!35}\underline{98.77} & \cellcolor{YellowOrange!35}\underline{84.65} \\

        \methodname & \cellcolor{ForestGreen!35}\textbf{96.94} & \cellcolor{ForestGreen!35}\textbf{75.36}  & \cellcolor{ForestGreen!35}\textbf{94.31} & \cellcolor{ForestGreen!35}\textbf{63.89} & \cellcolor{ForestGreen!35}\textbf{73.94} & \cellcolor{ForestGreen!35}\textbf{94.33} & \cellcolor{ForestGreen!35}\textbf{88.63} & \cellcolor{ForestGreen!35}\textbf{77.43} & 87.11 & \cellcolor{ForestGreen!35}\textbf{98.85} & \cellcolor{ForestGreen!35}\textbf{85.08} \\

        \hline
    \end{tabular}
    }
    \label{tab:16-shot}
    
    \vspace{-8pt}
\end{table*}

We evaluate our proposed method \methodname in comparison with state-of-the-art methods using ten benchmark datasets under two scenarios. We describe the experimental setup followed by a discussion of the main comparisons. Moreover, we ablate our main design choices regarding SAP and the PEFT adaptation and present both qualitative and quantitative results to show their effectiveness.

\noindent\textbf{Datasets.} We consider ten commonly used datasets for few-shot image classifier, namely, ImageNet~\cite{russakovsky2015imagenet} and Caltech101~\cite{fei2006one} for generic object classification, SUN397~\cite{xiao2010sun} for scene understanding, DTD~\cite{cimpoi14describing} for texture classification, FGVC Aircraft~\cite{maji13fine-grained}, Oxford Pets~\cite{parkhi12a}, Stanford Cars~\cite{krause20133d}, Food 101~\cite{bossard2014food} and Flowers 102~\cite{nilsback2008automated} for fine-grained classification and EuroSAT~\cite{helber2019eurosat} for satellite image classification.

\noindent\textbf{Evaluation protocol.} We evaluate our method in two scenarios. In the first (default) scenario, following~\cite{he2023synthetic}, we train and evaluate our model on all classes. In the second (base/new) scenario, we use half of the classes for training and evaluation (base classes) and the other half for evaluation only (new classes), following the split in~\cite{zhou2022conditional}. We report only the top-1 accuracy in the default scenario while, for the base/new scenario, we report the top-1 accuracy for the base classes (Base), the new classes (New), and their harmonic mean (H).

\noindent\textbf{Baseline methods.} In the default scenario, we compare our method against a list of PEFT methods including the Classifier Tuning from~\cite{he2023synthetic}, TPT (CoOp from~\cite{Zhou_2022});
VPT~\cite{jia2022visual}, and a combination of both VPT and TPT.
For the base/new scenario, we compare our method with the state-of-the-art methods including CoOp~\cite{Zhou_2022}, CoCoOp~\cite{zhou2022conditional}, MaPLe~\cite{Khattak_2023_CVPR} and LoGoPrompt~\cite{Shi_2023_ICCV}. We did not include Classifier Tuning as it consists of training a linear classifier that needs to know all classes a priori, thus not being applicable to unseen classes.

\noindent\textbf{Implementation details.} We build our method on top of CLIP~\cite{radford2021learning} with ViT-B/16 as vision encoder.
Our model is trained for $50$ epochs in all datasets, using AdamW~\cite{loshchilov2019decoupled} as optimizer, a cosine learning rate scheduler without warmup, and a weight decay of $ 1\times 10^{-3}$. Hyperparameters such as the learning rate and batch size are tuned on a per-dataset basis, as $lr \in \{2^{-i}, i \in [8, 15] \}$ and $bs \in \{16, 32, 64, 128, 256\}$.
We apply traditional data augmentation techniques, such as RandomResizedCrop, RandAugment~\cite{cubuk2019randaugment}, MixUp~\cite{yun2019cutmix}, CutMix~\cite{yun2019cutmix} and label smoothing~\cite{müller2020does}, depending on the dataset.
In experiments involving only real data, we set LoRA's $r=16$, $\alpha=32$, and $dropout=0.1$ for both the vision and text encoders, while for experiments with real and synthetic data, we set $r=64$ and $\alpha \in \{32, 64\}$ for the vision encoder depending on the dataset.
With the introduction of synthetic data during fine-tuning, we notice that a higher rank for the vision encoder can be beneficial as there is more visual data to leverage.
We set $\lambda \in \{0.5, 0.6, 0.7, 0.8\}$ when using synthetic data.
For SAP, we use LLaVA 1.5~\cite{liu2023improved} for captioning and a CLIP with ViT-H as vision encoder for filtering.
We choose as diffusion model a fine-tuned version of Stable Diffusion 1.5~\cite{rombach2022highresolution} on realistic images. As sampler, we use DPM-Solver++~\cite{lu2023dpmsolver} with 20 steps. We fixed the classifier-free guidance factor to 8. 
For the noising procedure, we perform a per-dataset choice of either 25\% or 75\% of the noising schedule. 
We set the number of synthetic samples per class $\numbersynthshots = 64$. More details are available in the Supplementary Material. 

\section{Results}

\begin{table*}[!ht]
    \centering
    \caption{16-shot results on all datasets for base/new classes averaged across 3 seeds.}
    \vspace{-8pt}
    \begin{subtable}[h]{0.45\textwidth}
    \centering
    \caption{Average across datasets}
    \resizebox{0.7\textwidth}{!}{
    \begin{tabular}{lcc|c}
        \hline
        & Base & New & H \\
        \hline        
        CLIP~\cite{radford2021learning} & 69.22 & 73.89 & 71.48 \\
        CoOp~\cite{Zhou_2022} & 82.02 & 63.94 & 72.08 \\
        CoCoOp~\cite{zhou2022conditional} & 80.28 & 71.51 & 75.65 \\
        MaPLe~\cite{Khattak_2023_CVPR} & 82.21 & \cellcolor{ForestGreen!35}
\textbf{74.75} & 78.33 \\
        LoGoPrompt~\cite{Shi_2023_ICCV} & 84.29 & 74.35 & \cellcolor{YellowOrange!35}\underline{79.02} \\
        \midrule
        \methodname w/o synth & \cellcolor{YellowOrange!35}\underline{85.16} & 74.07 & 78.95 \\
        \methodname & \cellcolor{ForestGreen!35}
\textbf{86.71}  & \cellcolor{YellowOrange!35}\underline{74.53} & \cellcolor{ForestGreen!35}
\textbf{79.74} \\
        \hline
    \end{tabular}
    }
    \label{tab:avg_base_new}
    \end{subtable}
    \hfill
    \begin{subtable}[h]{0.45\textwidth}
    \centering
    \caption{Caltech101}
    \resizebox{0.7\textwidth}{!}{
    \begin{tabular}{lcc|c}
        \hline
        & Base & New & H \\
        \hline
        CLIP~\cite{radford2021learning} & 96.84 & \cellcolor{YellowOrange!35}\underline{94.00} & 95.40 \\
        CoOp~\cite{Zhou_2022} & 98.00 & 89.81 & 93.73 \\
        CoCoOp~\cite{zhou2022conditional} & 97.96 & 93.81 & 95.84 \\
        MaPLe~\cite{Khattak_2023_CVPR} & 97.74 & \cellcolor{ForestGreen!35}
\textbf{94.36} & \cellcolor{YellowOrange!35}\underline{96.02} \\
        LoGoPrompt~\cite{Shi_2023_ICCV} & 98.19 & 93.78 & 95.93 \\
        \midrule
        \methodname w/o synth & \cellcolor{ForestGreen!35}
\textbf{98.58} & 92.98 & 95.70 \\
        \methodname & \cellcolor{YellowOrange!35}\underline{98.49}  & 93.85 & \cellcolor{ForestGreen!35}
\textbf{96.12} \\
        \hline
    \end{tabular}
    }
    \label{tab:caltech_base_new}
    \end{subtable}
    \hfill
    \begin{subtable}[h]{0.30\textwidth}
    \centering
    \caption{DTD}
    \resizebox{1.0\textwidth}{!}{
    \begin{tabular}{lcc|c}
        \hline
        & Base & New & H \\
        \hline
        CLIP~\cite{radford2021learning} & 53.24 & 59.90 & 56.37 \\
        CoOp~\cite{Zhou_2022} & 79.44 & 41.18 & 54.24 \\
        CoCoOp~\cite{zhou2022conditional} & 77.01 & 56.00 & 64.85 \\
        MaPLe~\cite{Khattak_2023_CVPR} & 80.36 & 59.18 & 68.16 \\
        LoGoPrompt~\cite{Shi_2023_ICCV} & \cellcolor{YellowOrange!35}\underline{82.87} & 60.14 & 69.70 \\
        \midrule
        \methodname w/o synth & 82.21 & \cellcolor{ForestGreen!35}
\textbf{65.62} & \cellcolor{ForestGreen!35}
\textbf{72.98} \\
        \methodname & \cellcolor{ForestGreen!35}
\textbf{83.57} & \cellcolor{YellowOrange!35}\underline{64.37} & \cellcolor{YellowOrange!35}\underline{72.70} \\
        \hline
    \end{tabular}
    }
    \label{tab:dtd_base_new}
    \end{subtable}
    \hfill
    \begin{subtable}[h]{0.30\textwidth}
    \centering
    \caption{EuroSAT}
    \resizebox{1.0\textwidth}{!}{
    \begin{tabular}{lcc|c}
        \hline
        & Base & New & H \\
        \hline
        CLIP~\cite{radford2021learning} & 56.48 & 64.05 & 60.03 \\
        CoOp~\cite{Zhou_2022} & 92.19 & 54.74 & 68.69 \\
        CoCoOp~\cite{zhou2022conditional} & 87.49 & 60.04 & 71.21 \\
        MaPLe~\cite{Khattak_2023_CVPR} & 94.07 & \cellcolor{YellowOrange!35}\underline{73.23} & 82.35 \\
        LoGoPrompt~\cite{Shi_2023_ICCV} & 93.67 & 69.44 & 79.75 \\
        \midrule
        \methodname w/o synth & \cellcolor{YellowOrange!35}\underline{97.72} & \cellcolor{ForestGreen!35}
\textbf{73.44} & \cellcolor{ForestGreen!35}
\textbf{83.86} \\
        \methodname & \cellcolor{ForestGreen!35}
\textbf{97.97} & 72.86 & \cellcolor{YellowOrange!35}\underline{83.53} \\
        \hline
    \end{tabular}
    }
    \label{tab:eurosat_base_new}
    \end{subtable}
    \hfill
    \begin{subtable}[h]{0.30\textwidth}
    \centering
    \caption{FGVC Aircraft}
    \resizebox{1.0\textwidth}{!}{
    \begin{tabular}{lcc|c}
        \hline
        & Base & New & H \\
        \hline
        CLIP~\cite{radford2021learning} & 27.19 & \cellcolor{ForestGreen!35}
\textbf{36.29} & 31.09 \\
        CoOp~\cite{Zhou_2022} & 40.44 & 22.30 & 28.75 \\
        CoCoOp~\cite{zhou2022conditional} & 33.41 & 23.71 & 27.74 \\
        MaPLe~\cite{Khattak_2023_CVPR} & 37.44 & \cellcolor{YellowOrange!35}\underline{35.61} & 36.50 \\
        LoGoPrompt~\cite{Shi_2023_ICCV} & 45.98 & 34.67 & \cellcolor{YellowOrange!35}\underline{39.53} \\
        \midrule
        \methodname w/o synth & \cellcolor{YellowOrange!35}\underline{48.50} & 32.23 & 38.71 \\
        \methodname & \cellcolor{ForestGreen!35}
\textbf{55.94} & 34.33 & \cellcolor{ForestGreen!35}
\textbf{42.53} \\
        \hline
    \end{tabular}
    }
    \label{tab:aircraft_base_new}
    \end{subtable}
    \hfill
    \begin{subtable}[h]{0.30\textwidth}
    \centering
    \caption{ImageNet}
    \resizebox{1.0\textwidth}{!}{
    \begin{tabular}{lcc|c}
        \hline
        & Base & New & H \\
        \hline
        CLIP~\cite{radford2021learning} & 72.43 & 68.14 & 70.22 \\
        CoOp~\cite{Zhou_2022} & 76.47 & 67.88 & 71.92 \\
        CoCoOp~\cite{zhou2022conditional} & 75.98 & 70.43 & 73.10 \\
        MaPLe~\cite{Khattak_2023_CVPR} & 76.66 & 70.54 & 73.47 \\
        LoGoPrompt~\cite{Shi_2023_ICCV} & 76.74 & \cellcolor{YellowOrange!35}\underline{70.83} & \cellcolor{YellowOrange!35}\underline{73.66} \\
        \midrule
        \methodname w/o synth & \cellcolor{YellowOrange!35}\underline{77.64} & 69.98 & 73.61 \\
        \methodname & \cellcolor{ForestGreen!35}
\textbf{78.34} & \cellcolor{ForestGreen!35}
\textbf{71.04} & \cellcolor{ForestGreen!35}
\textbf{74.51} \\
        \hline
    \end{tabular}
    }
    \label{tab:imagenet_base_new}
    \end{subtable}
    \hfill
    \begin{subtable}[h]{0.30\textwidth}
    \centering
    \caption{Oxford Pets}
    \resizebox{1.0\textwidth}{!}{
    \begin{tabular}{lcc|c}
        \hline
        & Base & New & H \\
        \hline
        CLIP~\cite{radford2021learning} & 91.17 & 97.26 & 94.12 \\
        CoOp~\cite{Zhou_2022} & 93.67 & 95.29 & 94.47 \\
        CoCoOp~\cite{zhou2022conditional} & 95.20 & 97.69 & 96.43 \\
        MaPLe~\cite{Khattak_2023_CVPR} & 95.43 & \cellcolor{ForestGreen!35}
\textbf{97.76} & \cellcolor{YellowOrange!35}\underline{96.58} \\
        LoGoPrompt~\cite{Shi_2023_ICCV} & 96.07 & 96.31 & 96.18 \\
        \midrule
        \methodname w/o synth & \cellcolor{YellowOrange!35}\underline{96.19} & 94.98 & 95.58 \\
        \methodname & \cellcolor{ForestGreen!35}
\textbf{96.40} & \cellcolor{YellowOrange!35}\underline{97.67}  & \cellcolor{ForestGreen!35}
\textbf{97.03} \\
        \hline
    \end{tabular}
    }
    \label{tab:pets_base_new}
    \end{subtable}
    \hfill
    \begin{subtable}[h]{0.30\textwidth}
    \centering
    \caption{Stanford Cars}
    \resizebox{1.0\textwidth}{!}{
    \begin{tabular}{lcc|c}
        \hline
        & Base & New & H \\
        \hline
        CLIP~\cite{radford2021learning} & 63.37 & \cellcolor{ForestGreen!35}
\textbf{74.89} & 68.65 \\
        CoOp~\cite{Zhou_2022} & 78.12 & 60.40 & 68.13 \\
        CoCoOp~\cite{zhou2022conditional} & 70.49 & 73.59 & 72.01 \\
        MaPLe~\cite{Khattak_2023_CVPR} & 72.94 & \cellcolor{YellowOrange!35}\underline{74.00} & 73.47 \\
        LoGoPrompt~\cite{Shi_2023_ICCV} & 78.36 & 72.39 & \cellcolor{YellowOrange!35}\underline{75.26} \\
        \midrule
        \methodname w/o synth & \cellcolor{YellowOrange!35}\underline{80.98} & 68.28 & 74.09 \\
        \methodname & \cellcolor{ForestGreen!35}
\textbf{84.07} & 68.75 & \cellcolor{ForestGreen!35}
\textbf{75.63} \\
        \hline
    \end{tabular}
    }
    \label{tab:cars_base_new}
    \end{subtable}
    \hfill
    \begin{subtable}[h]{0.30\textwidth}
    \centering
    \caption{SUN397}
    \resizebox{1.0\textwidth}{!}{
    \begin{tabular}{lcc|c}
        \hline
        & Base & New & H \\
        \hline
        CLIP~\cite{radford2021learning} & 69.36 & 75.35 & 72.23 \\
        CoOp~\cite{Zhou_2022} & 80.60 & 65.89 & 72.51 \\
        CoCoOp~\cite{zhou2022conditional} & 79.74 & 76.86 & 78.27 \\
        MaPLe~\cite{Khattak_2023_CVPR} & 80.82 & \cellcolor{ForestGreen!35}
\textbf{78.70} & 79.75 \\
        LoGoPrompt~\cite{Shi_2023_ICCV} & 81.20 & 78.12 & 79.63 \\
        \midrule
        \methodname w/o synth & \cellcolor{YellowOrange!35}\underline{81.87} & \cellcolor{YellowOrange!35}\underline{78.46} & \cellcolor{YellowOrange!35}\underline{80.13} \\
        \methodname & \cellcolor{ForestGreen!35}
\textbf{83.14} & 78.22 & \cellcolor{ForestGreen!35}
\textbf{80.61} \\
        \hline
    \end{tabular}
    }
    \label{tab:sun_base_new}
    \end{subtable}
    \hfill
    \begin{subtable}[h]{0.30\textwidth}
    \centering
    \caption{Food 101}
    \resizebox{1.0\textwidth}{!}{
    \begin{tabular}{lcc|c}
        \hline
        & Base & New & H \\
        \hline
        CLIP~\cite{radford2021learning} & 90.10 & 91.22 & 90.66 \\
        CoOp~\cite{Zhou_2022} & 88.33 & 82.26 & 85.19 \\
        CoCoOp~\cite{zhou2022conditional} & 90.70 & 91.29 & 90.99 \\
        MaPLe~\cite{Khattak_2023_CVPR} & 90.71 & \cellcolor{ForestGreen!35}
\textbf{92.05} & \cellcolor{ForestGreen!35}
\textbf{91.38} \\
        LoGoPrompt~\cite{Shi_2023_ICCV} & \cellcolor{YellowOrange!35}\underline{90.82} & 91.41 & \cellcolor{YellowOrange!35}\underline{91.11} \\
        \midrule
        \methodname w/o synth & \cellcolor{ForestGreen!35}
\textbf{91.00} & 90.51 & 90.75 \\
        \methodname & 90.56 & \cellcolor{YellowOrange!35}\underline{91.48} & 91.02 \\
        \hline
    \end{tabular}
    }
    \label{tab:food_base_new}
    \end{subtable}
    \hfill
    \begin{subtable}[h]{0.30\textwidth}
    \centering
    \caption{Flowers 102}
    \resizebox{1.0\textwidth}{!}{
    \begin{tabular}{lcc|c}
        \hline
        & Base & New & H \\
        \hline
        CLIP~\cite{radford2021learning} & 72.08 & \cellcolor{ForestGreen!35}
\textbf{77.80} & 74.83 \\
        CoOp~\cite{Zhou_2022} & 97.60 & 59.67 & 74.06 \\
        CoCoOp~\cite{zhou2022conditional} & 94.87 & 71.75 & 81.71 \\
        MaPLe~\cite{Khattak_2023_CVPR} & 95.92 & 72.46 & 82.56 \\
        LoGoPrompt~\cite{Shi_2023_ICCV} & \cellcolor{ForestGreen!35}
\textbf{99.05} & \cellcolor{YellowOrange!35}\underline{76.52} & \cellcolor{ForestGreen!35}
\textbf{86.34} \\
        \midrule
        \methodname w/o synth & 96.96 & 74.23 & \cellcolor{YellowOrange!35}\underline{84.09} \\
        \methodname & \cellcolor{YellowOrange!35}\underline{98.64} & 72.72 & 83.72 \\
        \hline
    \end{tabular}
    }
    \label{tab:flowers_base_new}
    \end{subtable}

\label{tab:base_new}
\vspace{-8pt}
\end{table*}

In Table \ref{tab:16-shot}, we present the results for the default scenario for all compared methods and ours with and without synthetic data.
First, we can see that TPT is better than VPT on average, although for EuroSAT, ImageNet, Oxford Pets, and Food 101, VPT outperforms TPT.
Nonetheless, \methodname significantly outperforms all compared methods, with and without synthetic data, on almost all datasets, except for Food 101 where using synthetic data reduces performance.
We suspect that this might be due to the limited generation capability in certain very fine-grained classes.
Although classifier tuning is competitive with \methodname in Caltech101, ImageNet, and Flowers 102, it is limited to only known classes, thus inappropriate for the base/new scenario, which involves unseen classes.

In addition, Table \ref{tab:base_new} presents the full results on the base/new scenario.
We can see that \methodname without synthetic data can improve the Base accuracy while being competitive with previous methods in New accuracy and H.
This shows that our fine-tuning strategy is effective without compromising the generalizability of the model to new classes.
Nonetheless, when adding our synthetic data, we further boost the base accuracy by +1.55\% on average while also improving New accuracy by +0.46\%, leading to a new state-of-the-art with an improvement of +0.72\% compared to the previous best-performing method LoGoPrompt~\cite{Shi_2023_ICCV} in H.

\subsection{Ablation}\label{sec:exp:ablation}

In this section, we demonstrate the effectiveness of our design choices both in the augmentation pipeline and the PEFT part.
For SAP, we ablate the effect of using the detail-rich captions obtained by $\mathcal{ICM}$ (LLaVA in our case), and the use of real samples as anchors during generation.
Additionally, we also demonstrate the effectiveness of the generated data when using it as augmentation with other fine-tuning methods, namely Classifier Tuning, VPT, TPT, and VPT + TPT.
For PEFT, we demonstrate that applying LoRA in both the image and text encoders of CLIP is important for reaching state-of-the-art results.
To demonstrate the effectiveness of our design choices, we select four representative datasets featuring different image recognition tasks: scene recognition (SUN397), fine-grained classes (FGVC Aircraft), satellite images (EuroSAT), and texture classification (DTD).

\noindent\textbf{How to generate the synthetic data?} We show the effect of using captions that are rich in semantic details and using real images as anchors for the generation. We ablate the individual contribution of each component in Table~\ref{abl:synth_components}. We can see that the optimal choice of which components to use is dataset-dependant, correlated with the nature of the data and the domain shift from the datasets diffusion models are trained on (\textit{e.g.} EuroSAT is a challenging domain to generate). On the other hand, we can see that when both components are used together, SAP is most beneficial for model generalization by bringing consistent improvement in all the datasets, leaving the amount of noise to inject as the only parameter to choose on a per-dataset basis.

\begin{table}[!htb]
\caption{Ablation of the SAP components.}
\vspace{-12pt}
\label{abl:synth_components}
\begin{center}
\resizebox{.42\textwidth}{!}{
\begin{tabular}{@{}cccccc@{}}
\toprule
\makecell{LLaVA\\caption} & \makecell{Real images\\guidance}      & EuroSAT & DTD & \makecell{FGVC\\Aircraft} & SUN397 \\ \midrule
\xmark              & \multicolumn{1}{c}{\xmark} &   93.98      &  74.43   & 63.64     &    74.40    \\
\xmark              & \multicolumn{1}{c}{\cmark} &   \cellcolor{ForestGreen!35}\textbf{94.36}      &  74.59   &   63.60   &   74.90     \\
\cmark              & \multicolumn{1}{c}{\xmark} &  94.08       &  \cellcolor{ForestGreen!35}\textbf{76.04}   &  \cellcolor{ForestGreen!35}\textbf{64.42}    &   \cellcolor{YellowOrange!35}\underline{75.22}     \\
\cmark              & \multicolumn{1}{c}{\cmark} &   \cellcolor{YellowOrange!35}\underline{94.31}      & \cellcolor{YellowOrange!35}\underline{75.36}    &  \cellcolor{YellowOrange!35}\underline{63.89}    &   \cellcolor{ForestGreen!35}\textbf{77.43}     \\ \bottomrule
\end{tabular}
}
\end{center}
\vspace{-10pt}
\end{table}

\noindent\textbf{Does SAP generalize to different fine-tuning methods?} We experiment with different tuning methods augmented with the synthetic data generated by SAP, demonstrating that SAP is generally effective, regardless of fine-tuning methods. We choose the same baselines as Table~\ref{tab:16-shot} and train them in the default scenario using the same data used for \methodname. From Table~\ref{abl:synth_baselines}, we see that all the baseline methods are able to achieve better results by including our synthetic data, proving that the images we generate are effective regardless of the method used to fine-tune CLIP. The only notable exception is classifier tuning, where the use of synthetic data for FGVC Aircraft marginally decreases the performance. We deem this result to be the saturation of the learnable parameters of the classifier and the difficulty of this recognition task, which makes adding more data non-beneficial given the low-parameter regime.

\begin{table}[!htb]
\caption{Effect of using synthetic data on the baselines in the default scenario.}
\vspace{-8pt}
\label{abl:synth_baselines}
\begin{center}
\resizebox{.45\textwidth}{!}{
\begin{tabular}{@{}lcccc@{}}
\toprule
            & EuroSAT & DTD & \makecell{FGVC\\Aircraft} & SUN397 \\ \midrule
\multicolumn{1}{l}{Classifier Tuning~\cite{he2023synthetic}}          &   87.13      &  73.64   &  \cellcolor{ForestGreen!35}\textbf{46.73}    & 76.16       \\ 
\multicolumn{1}{l}{ \textcolor{gray}{+synthetic} }          &  \cellcolor{ForestGreen!35}\textbf{88.14}       & \cellcolor{ForestGreen!35}\textbf{73.78}    &  46.47    &  \cellcolor{ForestGreen!35}\textbf{76.90}      \\ \midrule

\multicolumn{1}{l}{Visual Prompt Tuning~\cite{jia2022visual}}          &     92.33    &  66.06   &  36.21    & 70.47       \\ 
\multicolumn{1}{l}{\textcolor{gray}{+ synthetic}} &  \cellcolor{ForestGreen!35}\textbf{93.01}       &   \cellcolor{ForestGreen!35}\textbf{68.62}  &   \cellcolor{ForestGreen!35}\textbf{39.86}   & \cellcolor{ForestGreen!35}\textbf{71.70}       \\ \midrule
\multicolumn{1}{l}{Text Prompt Tuning~\cite{Zhou_2022}} &         
 \cellcolor{ForestGreen!35}\textbf{87.05} & 70.73    &   45.50   &   72.95     \\
\multicolumn{1}{l}{\textcolor{gray}{+ synthetic}} &   86.68      &  \cellcolor{ForestGreen!35}\textbf{71.41}  &  \cellcolor{ForestGreen!35}\textbf{46.81}    & \cellcolor{ForestGreen!35}\textbf{73.90}       \\ \midrule
\multicolumn{1}{l}{VPT + TPT~\cite{jia2022visual,Zhou_2022}} &       90.42  &   72.02  &   48.03   &  72.58      \\
\multicolumn{1}{l}{\textcolor{gray}{+ synthetic}} &   \cellcolor{ForestGreen!35}\textbf{92.88}      &  \cellcolor{ForestGreen!35}\textbf{72.62}   &  \cellcolor{ForestGreen!35}\textbf{50.55}    &   \cellcolor{ForestGreen!35}\textbf{73.79}     \\ \midrule
\multicolumn{1}{l}{Ours (LoRA~\cite{hu2021lora})}        &   94.25      &  74.31   &  62.09    &  77.01    \\ 
\multicolumn{1}{l}{\textcolor{gray}{+ synthetic (\methodname)}} &    \cellcolor{ForestGreen!35}\textbf{94.31}     &  \cellcolor{ForestGreen!35}\textbf{75.36}   &  \cellcolor{ForestGreen!35}\textbf{63.89}    &    \cellcolor{ForestGreen!35}\textbf{77.43}    \\ \bottomrule
\end{tabular}
}
\end{center}
\vspace{-15pt}
\end{table}

\noindent\textbf{Which encoder should we adapt?} We also ablate the effect of adding LoRA only to the vision encoder, only the text encoder, or both. As shown in Table \ref{abl:lora}, we can see that depending on the dataset, there are different effects of adapting only one of the modalities.
This depends mostly on how large the domain shift is for each modality \wrt the pre-training datasets of CLIP.
For example, in datasets such as EuroSAT, where the text modality is already well-separated by CLIP but the visual aspects of the data greatly differ from the pre-training data, adapting only the vision encoder is enough to bring results close to our best.
It is also interesting to see that adding LoRA only to the vision encoder is a better choice than adding it only to the text encoder.
This is different from adding learnable prompts, as we saw that TPT achieved better results than VPT.
When using LoRA, we can properly adapt the vision encoder, which results in higher gains than adapting only the text encoder.
Nonetheless, both encoders are complementary, therefore adapting the text encoder contributes to further improvements.

\begin{table}[!htb]
\caption{Effect of LoRA for adaptation.}
\vspace{-8pt}
\label{abl:lora}
\begin{center}
\resizebox{.37\textwidth}{!}{
\begin{tabular}{l|cccc}
\toprule
        LoRA & DTD & EuroSAT & \makecell{FGVC\\Aircraft} & SUN397 \\
\midrule
Vision Only & 72.79 & \cellcolor{ForestGreen!35}\textbf{93.88} & 60.03 & 72.76 \\
          \textcolor{gray}{+ synthetic}               & \cellcolor{ForestGreen!35}\textbf{73.78} & 93.30 & \cellcolor{ForestGreen!35}\textbf{62.50} & \cellcolor{ForestGreen!35}\textbf{75.18} \\
\midrule
Text Only & 70.92 & \cellcolor{ForestGreen!35}\textbf{86.97} & 46.68 & \cellcolor{ForestGreen!35}\textbf{76.04} \\
           \textcolor{gray}{+ synthetic}          & \cellcolor{ForestGreen!35}\textbf{71.93} & 86.91 & \cellcolor{ForestGreen!35}\textbf{47.06} & 75.90 \\
\midrule
Both (Ours) & 74.31 & 94.25 & 62.09 & 77.01 \\
             \textcolor{gray}{+ synthetic}        & \cellcolor{ForestGreen!35}\textbf{75.36} & \cellcolor{ForestGreen!35}\textbf{94.31} & \cellcolor{ForestGreen!35}\textbf{63.89} & \cellcolor{ForestGreen!35}\textbf{77.43} \\
\bottomrule                     
\end{tabular}
}
\end{center}
\vspace{-15pt}
\end{table}

\subsection{Qualitative results}
\begin{figure}
\begin{center}

\setlength\extrarowheight{-10pt}
\setlength\tabcolsep{1.pt} 

\begin{tabular}{cccc}
\rotatebox[origin=c]{90}{He et al.~\cite{he2023synthetic}} & \includegraphics[width=.21\linewidth,valign=m]{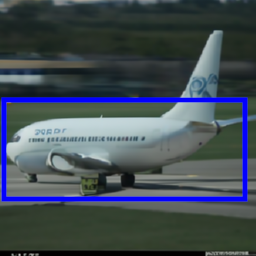} & \includegraphics[width=.21\linewidth,valign=m]{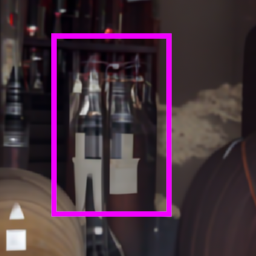} & \includegraphics[width=.21\linewidth,valign=m]{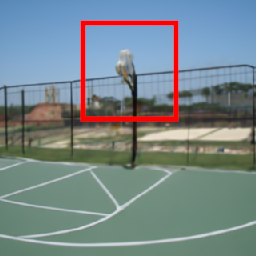} \\\\

\rotatebox[origin=c]{90}{\makecell{Stable\\Diffusion~\cite{rombach2022highresolution}}} & 
\includegraphics[width=.21\linewidth,valign=m]{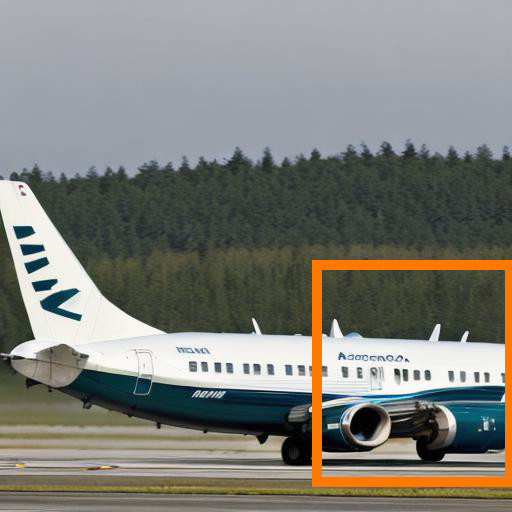} & 
\includegraphics[width=.21\linewidth,valign=m]{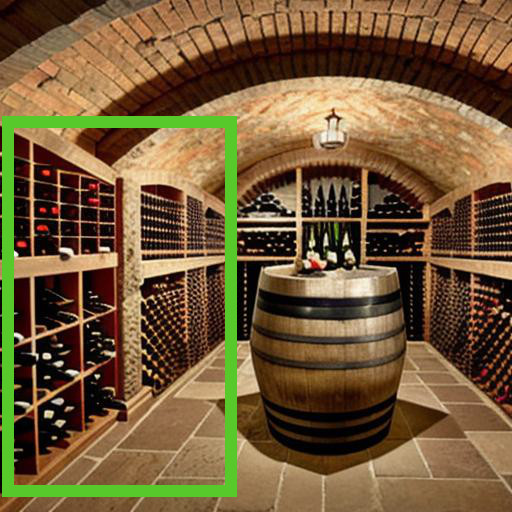} &
\includegraphics[width=.21\linewidth,valign=m]{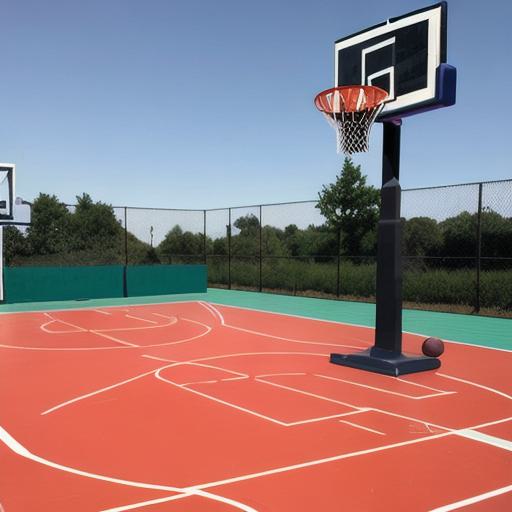}  \\\\

\rotatebox[origin=c]{90}{\makecell{SAP\\(Full)}} & 
\includegraphics[width=.21\linewidth,valign=m]{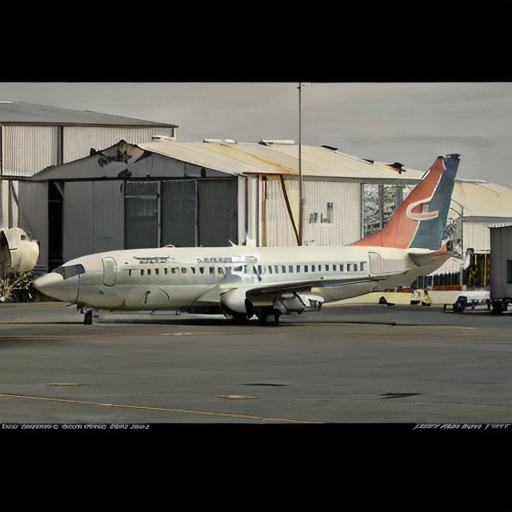} & \includegraphics[width=.21\linewidth,valign=m]{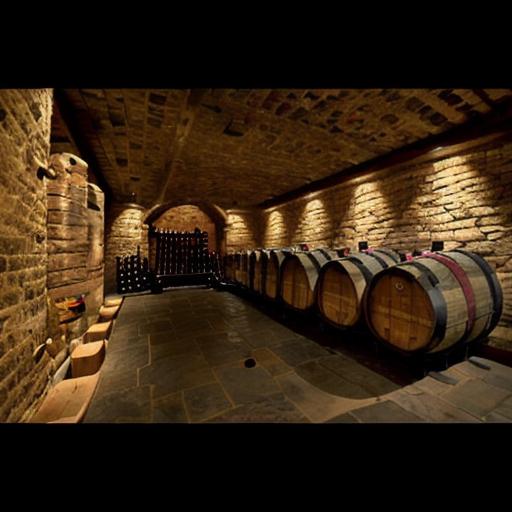} & \includegraphics[width=.21\linewidth,valign=m]{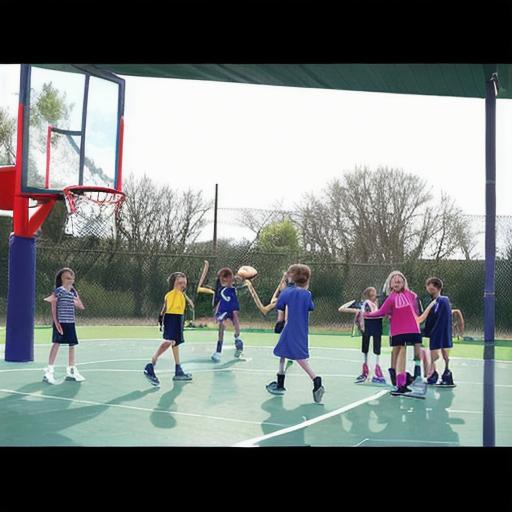}

\end{tabular}

\end{center}
\vspace{-10pt}
\caption{He \textit{et al.}~\cite{he2023synthetic} generation pipeline compared with our SAP and Stable Diffusion~\cite{rombach2022highresolution}. The generated classes are, from left to right, "a Boeing 737-200", "a wine cellar barrel storage" and "an outdoor basketball court". Images from~\cite{he2023synthetic} exhibit \textcolor{RoyalBlue}{wrong proportions}, a general \textcolor{Magenta}{lack of details}, or \textcolor{Red}{missing fundamental objects}. Images naively generated from scratch with a generic prompt exhibit \textcolor{Orange}{cropped objects} or \textcolor{LimeGreen}{wrong semantics}. Our SAP, which leverages real data and rich captions, does not exhibit such drawbacks.} 
\vspace{-6pt}
\label{fig:qualitatives}

\end{figure}

We present some generated samples in Figure~\ref{fig:qualitatives} by our SAP with both the semantically rich captions and the guidance from the real samples in $\supportset$, by GLIDE in \cite{he2023synthetic} and a Stable Diffusion~\cite{rombach2022highresolution} model with standard CLIP prompts as input. 
We can see that the samples generated by GLIDE in \cite{he2023synthetic} generally present lower quality with fewer details, \eg they miss the basket in the basketball court in the third column. On the other hand, generation from scratch with a generic prompt (second row) leads to generally good-looking samples, however also lacks details, \eg the missing front part of the airplane or the general absence of \textit{barrels} in the \textit{barrel storage}. Instead, our SAP generation (the last row) exhibits a higher degree of detail with the presence of the correct and complete object in the images.

\section{Conclusion} \label{sec:conclusion}
We presented \methodname, a novel method for few-shot learning with two main contributions: a novel design of synthetic augmentation pipeline for in-domain diversity and a novel application of a parameter-efficient fine-tuning method to VLMs for effective adaptation.
For the synthetic data generation, we propose to apply noise to the real samples and use them as a starting point for the diffusion process.
Additionally, we leverage textual information, in the form of captions, to enrich the details of the generated data.
Finally, we leveraged LoRA for fine-tuning both the vision and the text encoder, allowing us to effectively adapt a pre-trained VLM with limited samples.
Our method was evaluated against previous approaches in two scenarios on ten benchmark datasets, achieving the new state-of-the-art in few-shot image classification.

\noindent\textbf{Limitations.} Using a pre-trained diffusion model as the generative backbone might limit the semantics of the generated images for domains on which the model was not explicitly trained, \textit{e.g.} medical images. Furthermore, since we are adapting a pre-trained CLIP model, if the domain between the pre-training data and the few-shot data differs too much, fine-tuning might not suffice.

\noindent\textbf{Broader Societal Impacts.} 
Although few-shot image classification has already alleviated the need for massive training data, data privacy still remains a concern in case the support set contains sensitive information. The potential biases can be exacerbated by limited training samples, leading to unfair outcomes.

{
    \small
    \bibliographystyle{ieeenat_fullname}
    \bibliography{main}
}

\clearpage
\setcounter{page}{1}
\maketitlesupplementary
\appendix

In the supplementary material, we provide additional results.
In Section~\ref{sec:abl:less_synthetic}, we provide an ablation on the number of synthetic images used during training. In Section~\ref{sec:abl:fewer_shots}, we show the effect of using less real samples for \methodname and all the baselines in the default scenario.
In Section~\ref{sec:abl:parameters}, we provide a complete overview of all the parameters used for our \methodname for all the components, divided by dataset.
Finally, in Section~\ref{sec:abl:more_quali}, we show additional results generated by our SAP, proving the effectiveness of our augmentation pipeline.

\section{Effect of using less synthetic images}
\label{sec:abl:less_synthetic}
We study the effect of the number of synthetic data used in the fine-tuning process.
In Table~\ref{abl:synth_samples}, we present these results on the four representative datasets for our ablation studies. 
We adopt the 16-shot default scenario and vary the number of synthetic images from 4 to 64 (our default value).
We can see that a higher number of synthetic images leads to better performance in all the datasets.
Nonetheless, we also notice that the performance achieved by 64 synthetic samples is only marginally better than the ones obtained by 32 synthetic samples on all datasets, indicating that performance is saturating at 64.
Such saturation may be due to the limited amount of diversity the 16 original samples can provide for generating synthetic data.
Additionally, there is an increased computational burden on generating a significantly higher number of synthetic images.

\section{Fewer shots}
\label{sec:abl:fewer_shots}
\begin{figure*}[!htbp]
    \centering
    \begin{tabular}{cc}
        EuroSAT & DTD \\
        \includegraphics[width=0.35\textwidth]{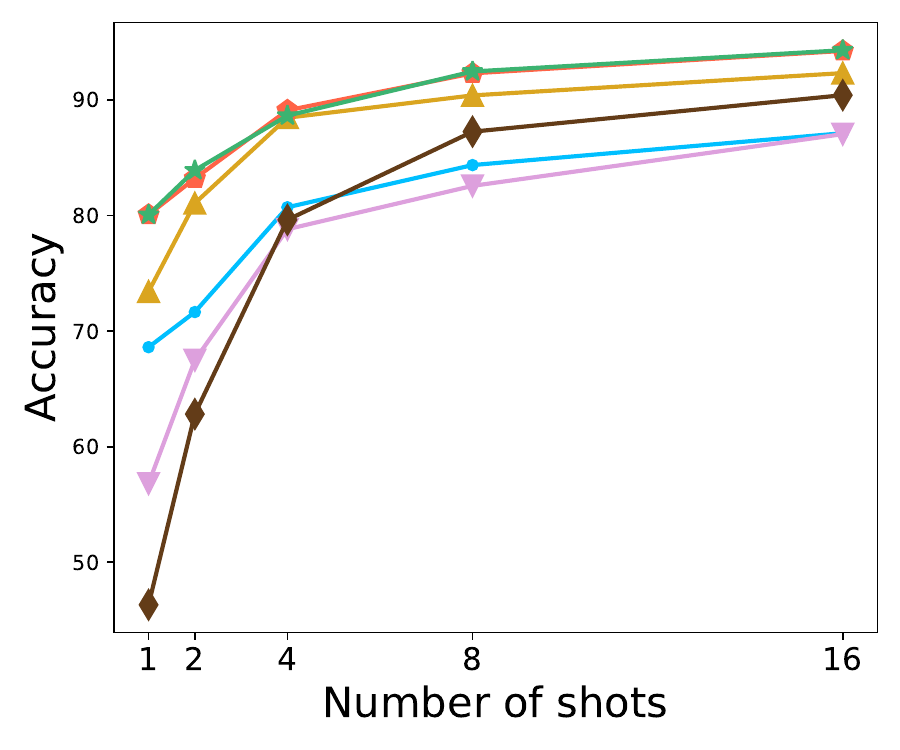} &
        \includegraphics[width=0.35\textwidth]{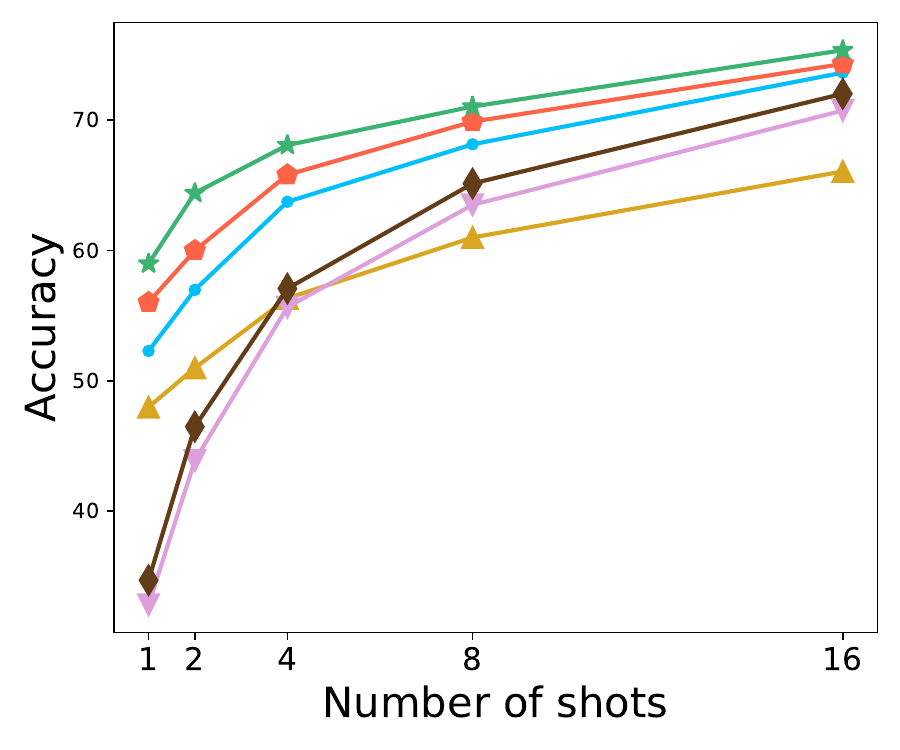} \\

        FGVC Aircraft & SUN397 \\
        \includegraphics[width=0.35\textwidth]{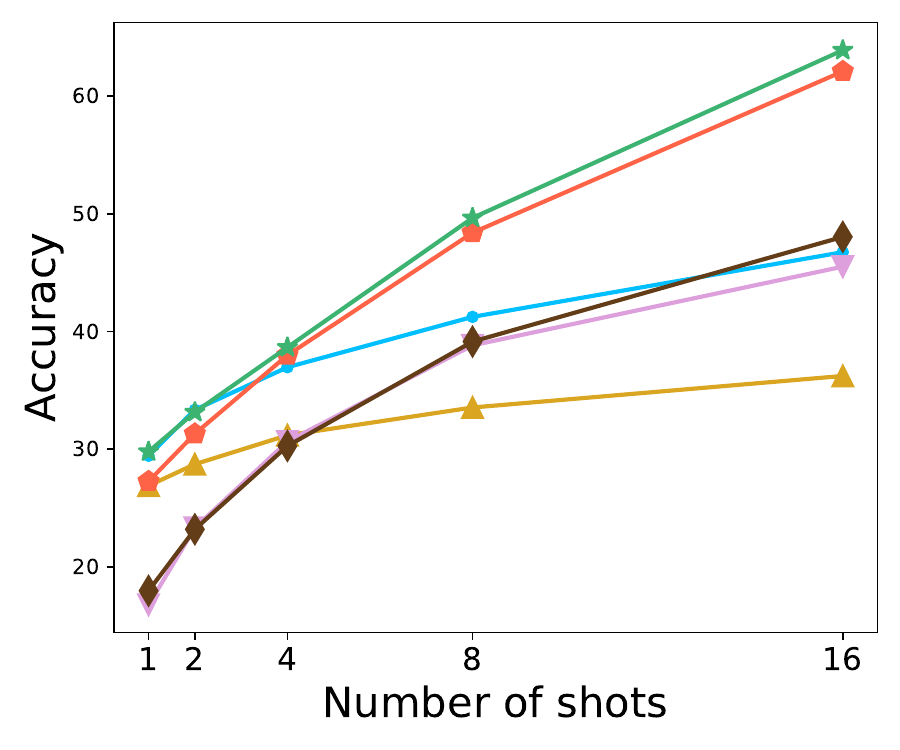} & \includegraphics[width=0.35\textwidth]{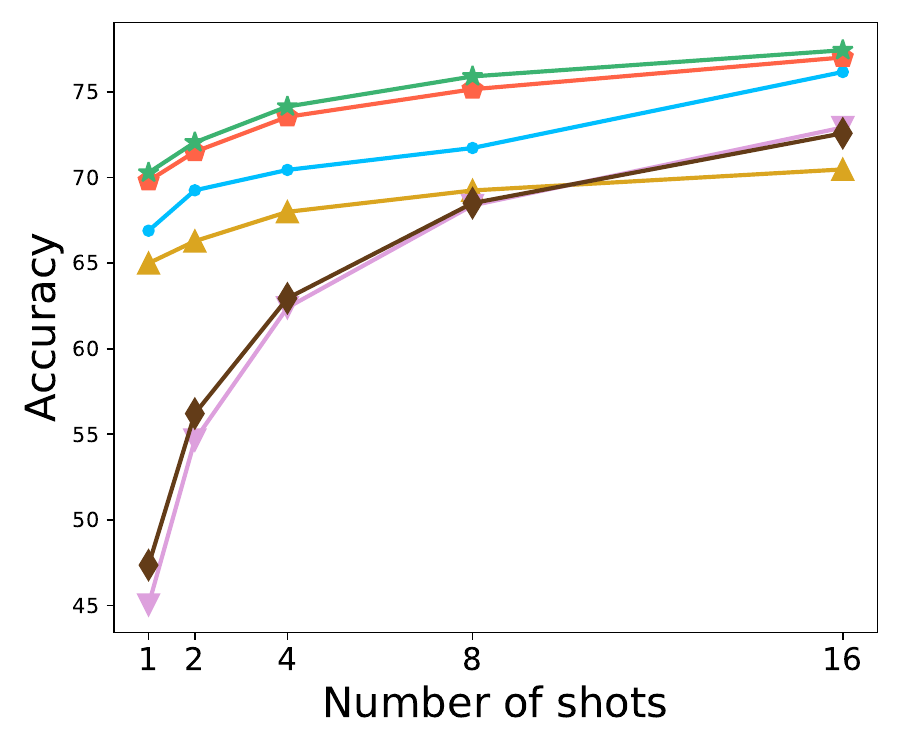} \\
    \end{tabular}
    \includegraphics[width=0.9\textwidth]{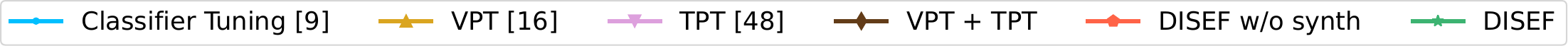}
    \caption{Fewer shots for the default scenario for Classifier Tuning \cite{he2023synthetic}, VPT \cite{jia2022visual}, TPT \cite{Zhou_2022}, VPT + TPT, \methodname w/o synthetic data and \methodname.}
    \label{fig:fewer-shot}
\end{figure*}

In this section, we ablate the effect of training with fewer shots, \ie, fewer real images per class. More specifically, we evaluate the baseline methods, classifier tuning \cite{he2023synthetic}, VPT \cite{jia2022visual}, TPT \cite{Zhou_2022} and VPT+TPT, on the default scenario, using a ViT-B/16 as vision encoder. We consider the same four ablation datasets, EuroSAT, DTD, FGVC Aircraft, and SUN397, and conduct experiments with 1, 2, 4, and 8 shots.
Since our synthetic data generation is conditioned on the images and their captions, for each fewer-shot experiment, we re-generate the data such that only those images are used for conditioning, and we still generate 64 synthetic images.
We present these results in Figure \ref{fig:fewer-shot}.
First, it is interesting to see that VPT performs much better than TPT with less real data in DTD, FGVC Aircraft, and SUN397, although it converges to a much lower value.
Nonetheless, we can see that \methodname without synthetic data performs better than all methods regardless of the number of shots.
When coupled with synthetic data, our method improves on all datasets and with different shots, with the exception of EuroSAT for 4-shot.
Even in the most extreme case 1-shot where we use only one image and its caption as a starting point, our generation is still powerful and versatile enough to achieve the most competitive accuracy.

\begin{table}[!t]
\caption{Effect of reducing the number of synthetic images.}
\begin{center}
\resizebox{.45\textwidth}{!}{
\begin{tabular}{@{}lcccc@{}}
\toprule
Number of synthetic shots & EuroSAT & DTD   & FGVC Aircraft & SUN397 \\ \midrule
4                         & 92.42   & 69.98 &    57.69           &   76.08     \\
8                         & 92.90   & 73.07 &   61.93            &  77.14      \\
16                        & 94.16   & 74.05 &    63.33           &  77.39      \\
32                        & 94.13   & 75.00 &       63.79        &  77.41      \\
64 (ours)                 & \textbf{94.31}   & \textbf{75.36} & \textbf{63.89}         & \textbf{77.43}  \\ \bottomrule
\end{tabular}
}
\end{center}
\label{abl:synth_samples}

\vspace{-0.5cm}
\end{table}

\section{Hyperparameters per dataset}
\label{sec:abl:parameters}

We perform a grid search on the hyperparameters of both the fine-tuning and generation process on a per-dataset basis and present these values in Table \ref{tab:parameters}.
For the fine-tuning hyperparameters, we experimented with different batch sizes, learning rates, LoRA's $r$ and $\alpha$ for the vision encoder, $\lambda$, cutmix, mixup and label-smoothing.
Batch size and learning rate were searched in the ranges of $\{16, 32, 64, 128, 256\}$ and $\{2^{-i}, i \in [8, 15]\}$.
For CutMix, MixUp and label-smoothing we used either $[0.0, 0.0, 0.0]$, $[0.1, 0.1, 0.1]$ or $[0.8, 1.0, 0.1]$ for each respective parameter.
LoRA's $r$ and $\alpha$ for the vision encoder were searched in the ranges of $\{16, 32, 64\}$ and $\{32, 64\}$.
$\text{Augmentation} = \text{True}$  means using RandAugment \cite{cubuk2019randaugment} with default parameters followed by RandomResizedCrop, from the official Pytorch implementation.
For the synthetic parameters, we search only for the step to start generation from, in $\{5, 15\}$, with $5$ meaning an initial sample closer to Gaussian noise, while $15$ implies a sample closer to the initial image.

\section{Synthetic samples generated by SAP}
\label{sec:abl:more_quali}

\begin{figure*}[!ht]
\caption{Qualitative example of diverse in-domain synthetis of our SAP. In \fcolorbox{red}{white}{red box} samples from the ground truth 16 shots, below different synthesis results from our model.}
\centering

\setlength{\tabcolsep}{0.em} 
\renewcommand{\arraystretch}{0.}

\fboxsep=0mm
\fboxrule=3pt

\begin{tabular}{@{}cccc@{}}
\toprule
EuroSAT & DTD    & FGVC Aircraft & SUN397      \\ \midrule
AnnualCrop  & Banded & Supermarine Spitfire        & Bar         \\
       \fcolorbox{red}{yellow}{\includegraphics[width=0.10\linewidth]{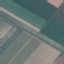}} & \fcolorbox{red}{yellow}{\includegraphics[width=0.10\linewidth]{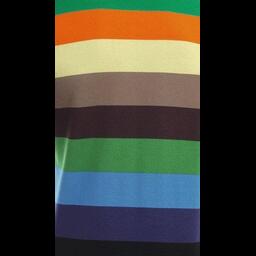}}       &        \fcolorbox{red}{yellow}{\includegraphics[width=0.10\linewidth]{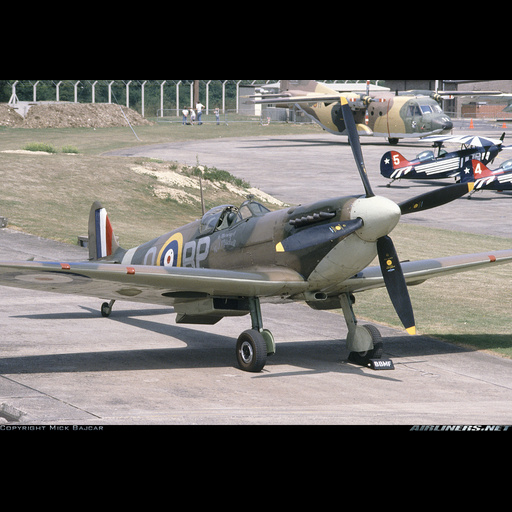}}       & \fcolorbox{red}{yellow}{\includegraphics[width=0.10\linewidth]{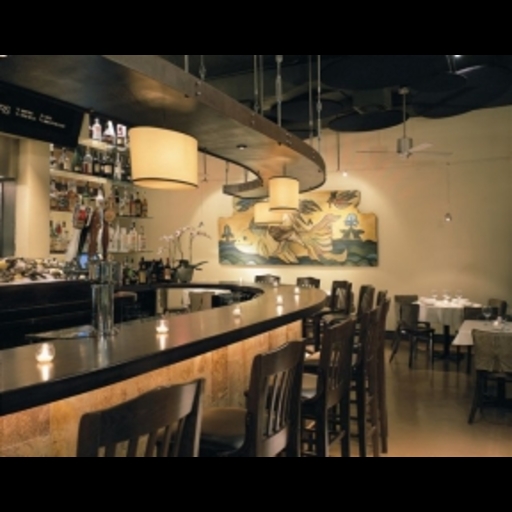}}             \\
        \includegraphics[width=0.10\linewidth]{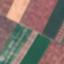} &  \includegraphics[width=0.10\linewidth]{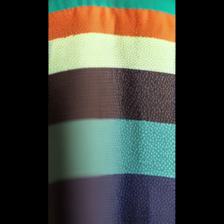}      &   \includegraphics[width=0.10\linewidth]{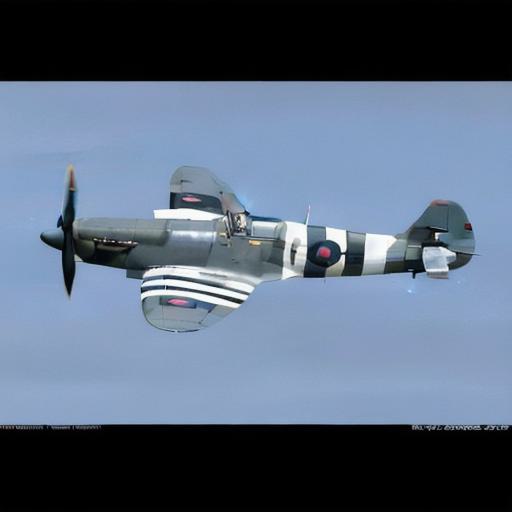}            & \includegraphics[width=0.10\linewidth]{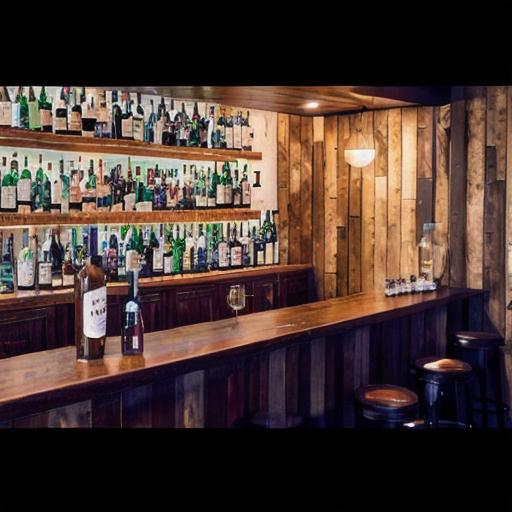}            \\
        \includegraphics[width=0.10\linewidth]{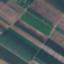} &  \includegraphics[width=0.10\linewidth]{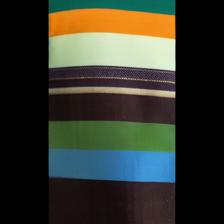}      &   \includegraphics[width=0.10\linewidth]{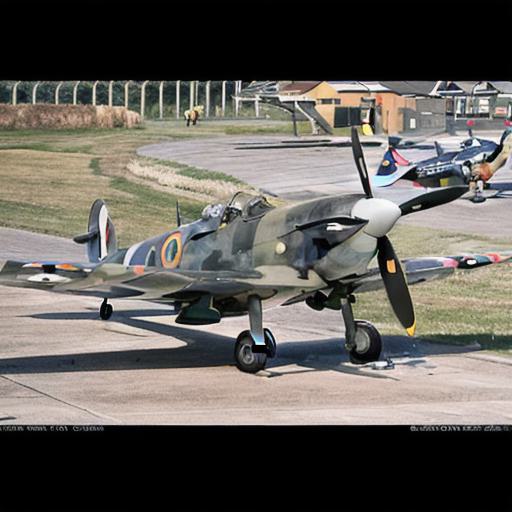}            &  \includegraphics[width=0.10\linewidth]{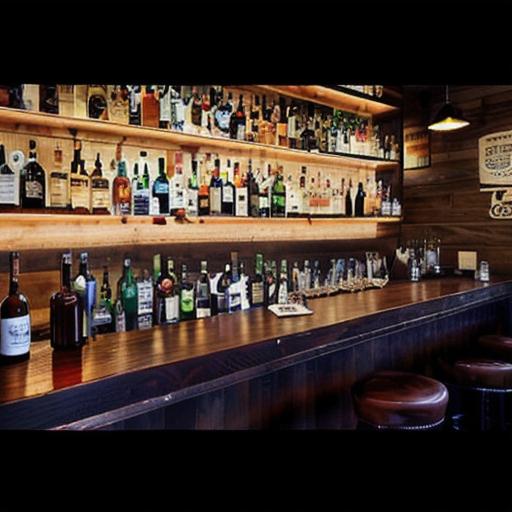}           \\
        \includegraphics[width=0.10\linewidth]{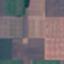} &   \includegraphics[width=0.10\linewidth]{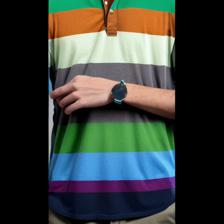}     &   \includegraphics[width=0.10\linewidth]{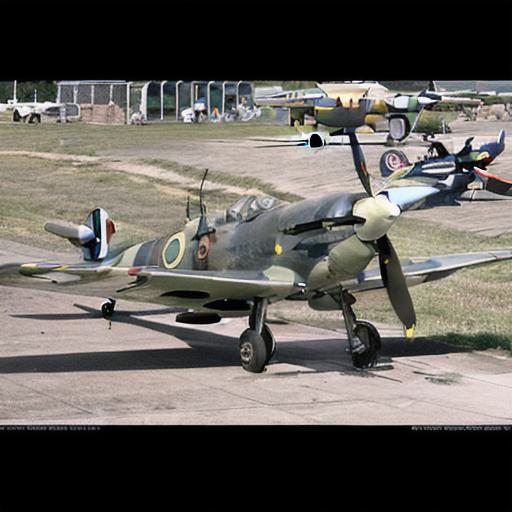}            &  \includegraphics[width=0.10\linewidth]{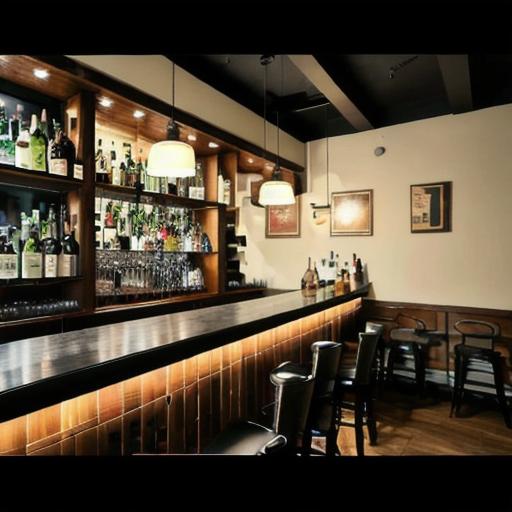}           \\
        \includegraphics[width=0.10\linewidth]{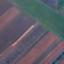} & \includegraphics[width=0.10\linewidth]{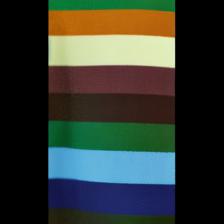}       & \includegraphics[width=0.10\linewidth]{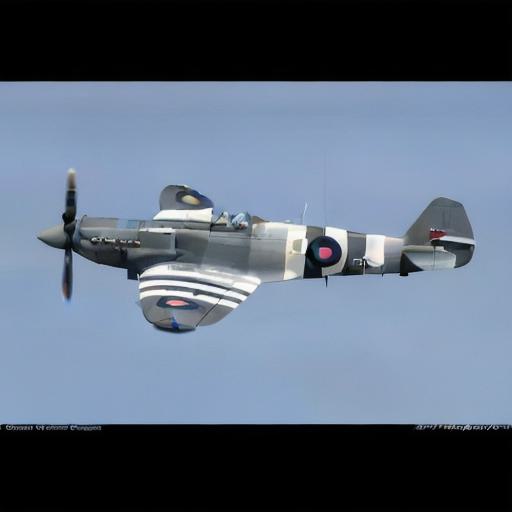}              &  \includegraphics[width=0.10\linewidth]{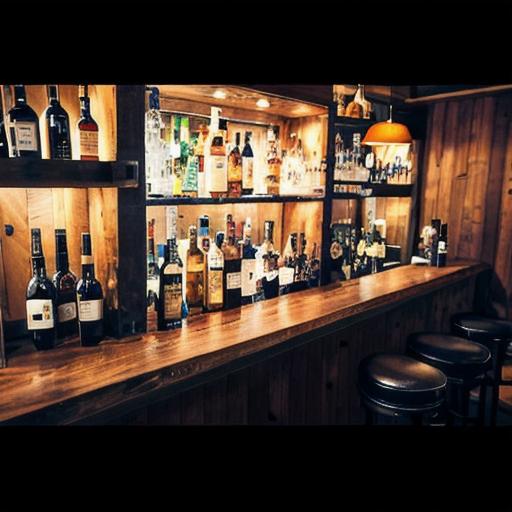}           \\ \bottomrule
\end{tabular}
\label{fig:another_one}
\end{figure*}

\begin{figure*}
\begin{center}
\caption{Randomly selected images generated by our SAP for the ten datasets. Class names can be found in Table~\ref{tab:qualitatives_classnames}.}

\label{fig:more_qualitatives}

\setlength{\tabcolsep}{0.em} 
\renewcommand{\arraystretch}{0.}

\begin{tabular}{@{}cccccccccc@{}}
\toprule

Caltech101 & DTD & EuroSAT & FGVCA & ImageNet & OxfordPets & Cars & SUN397 & Food 101 & Flowers102 \\ \midrule


\includegraphics[width=0.1\textwidth]{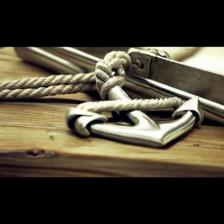} & \includegraphics[width=0.1\textwidth]{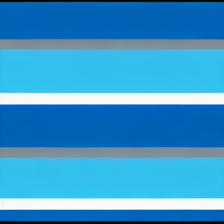} & \includegraphics[width=0.1\textwidth]{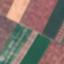}  & \includegraphics[width=0.1\textwidth]{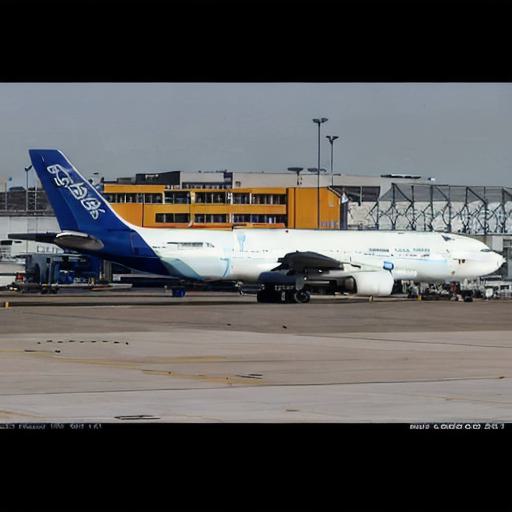} & \includegraphics[width=0.1\textwidth]{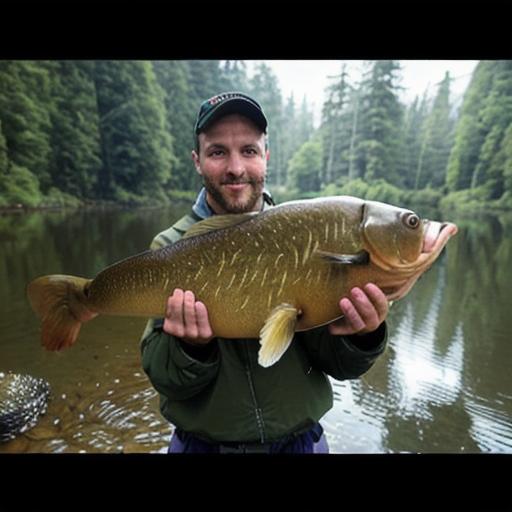} & \includegraphics[width=0.1\textwidth]{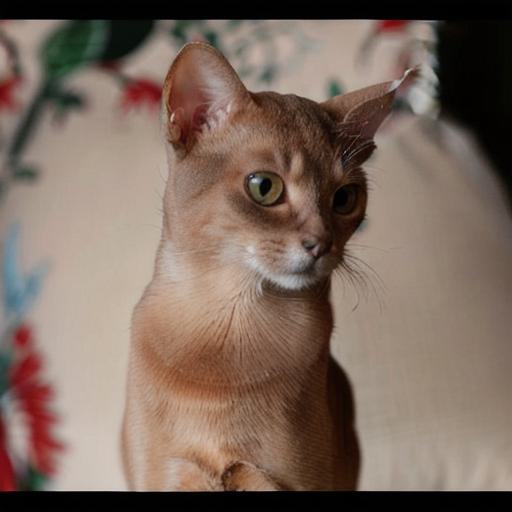} & \includegraphics[width=0.1\textwidth]{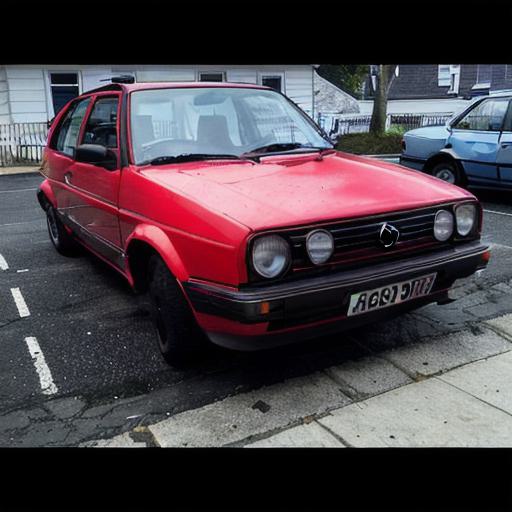} & \includegraphics[width=0.1\textwidth]{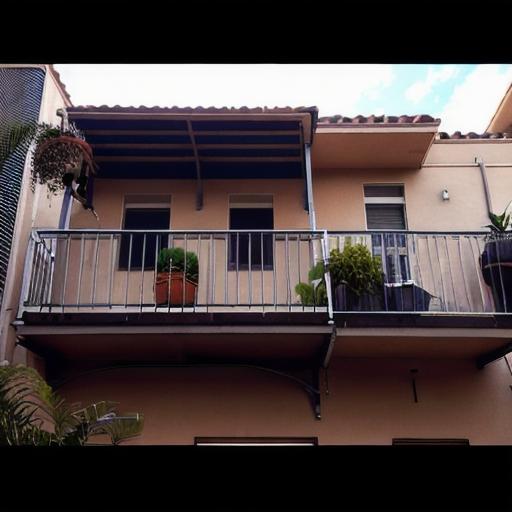} & \includegraphics[width=0.1\textwidth]{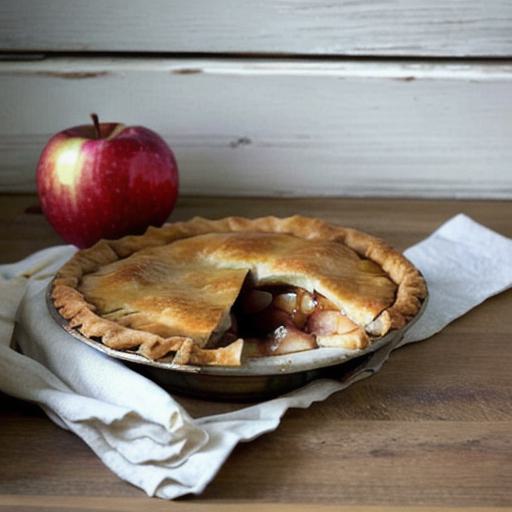}  & \includegraphics[width=0.1\textwidth]{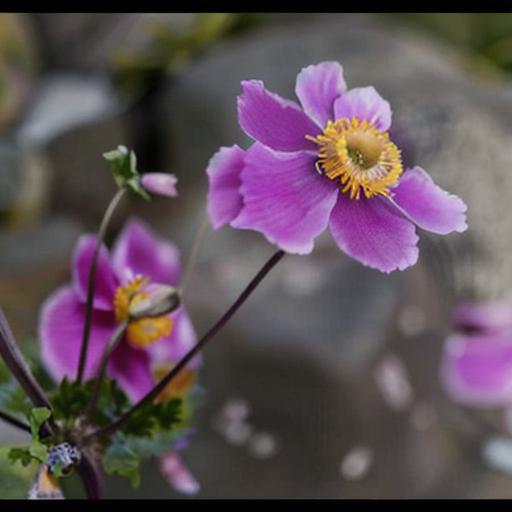} \\
\includegraphics[width=0.1\textwidth]{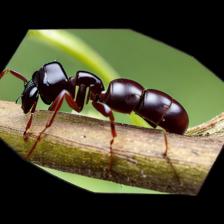} & \includegraphics[width=0.1\textwidth]{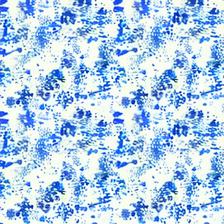} & \includegraphics[width=0.1\textwidth]{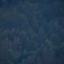}  & \includegraphics[width=0.1\textwidth]{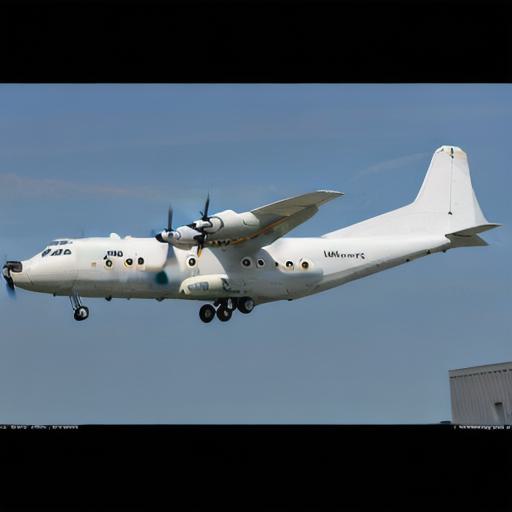} & \includegraphics[width=0.1\textwidth]{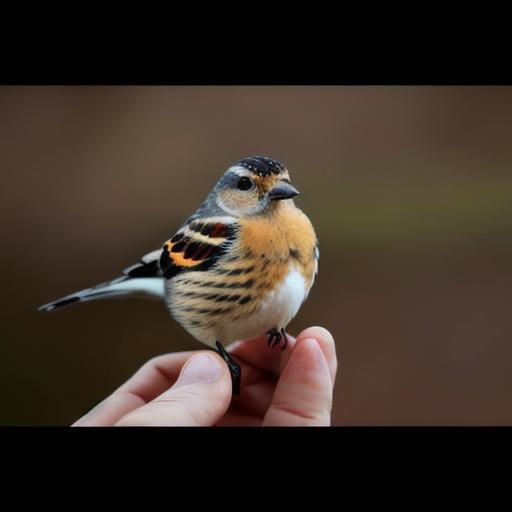} & \includegraphics[width=0.1\textwidth]{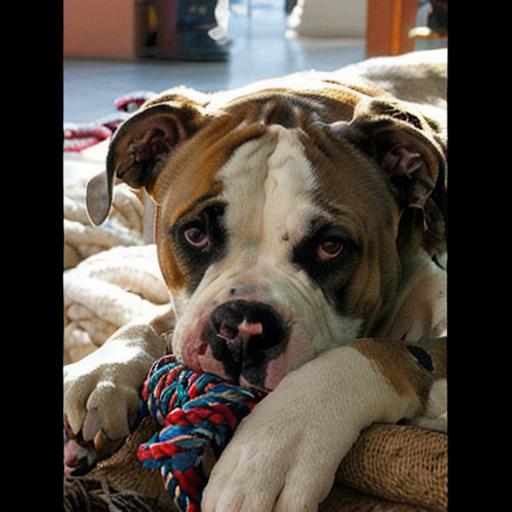} & \includegraphics[width=0.1\textwidth]{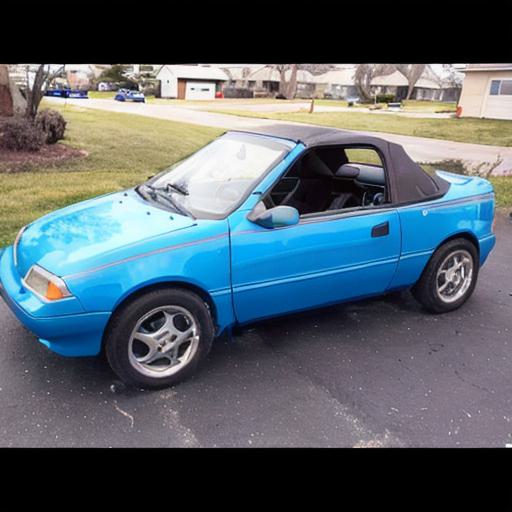} & \includegraphics[width=0.1\textwidth]{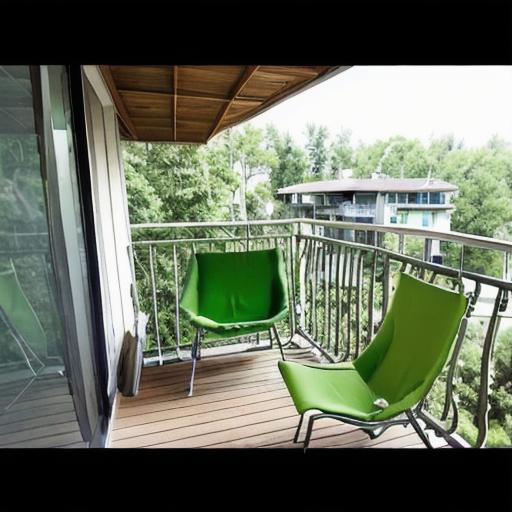} & \includegraphics[width=0.1\textwidth]{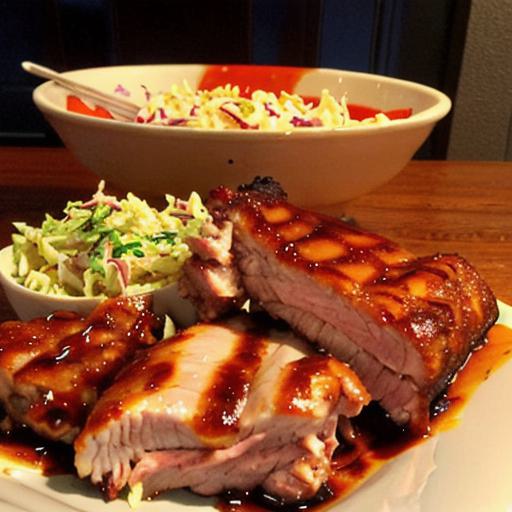} & \includegraphics[width=0.1\textwidth]{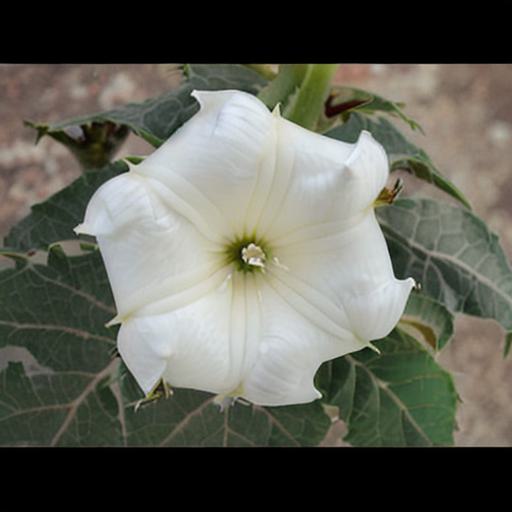} \\
\includegraphics[width=0.1\textwidth]{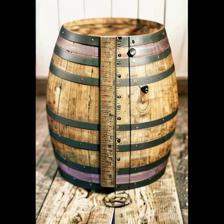} & \includegraphics[width=0.1\textwidth]{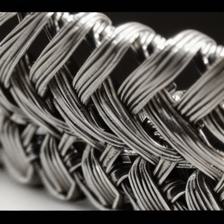} & \includegraphics[width=0.1\textwidth]{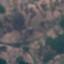} & \includegraphics[width=0.1\textwidth]{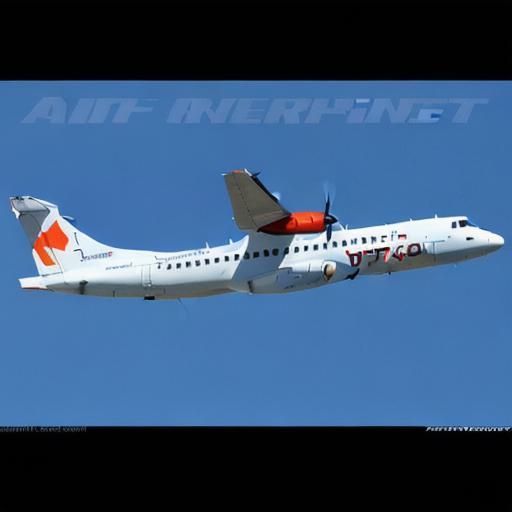} & \includegraphics[width=0.1\textwidth]{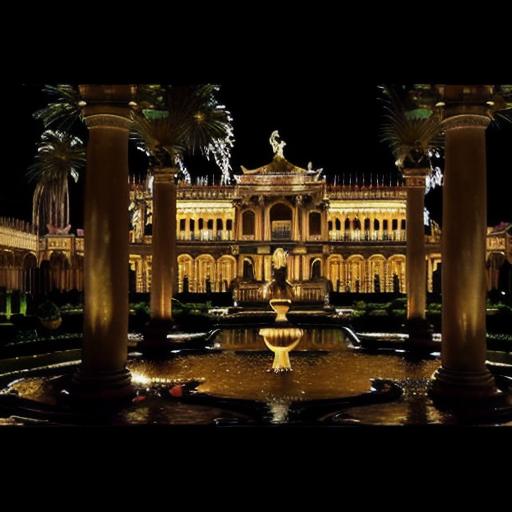} & \includegraphics[width=0.1\textwidth]{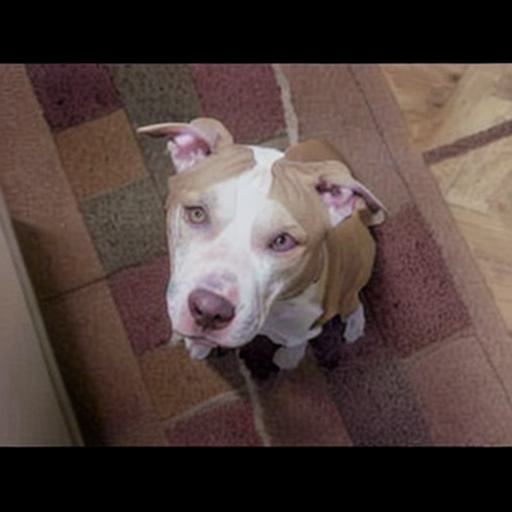} & \includegraphics[width=0.1\textwidth]{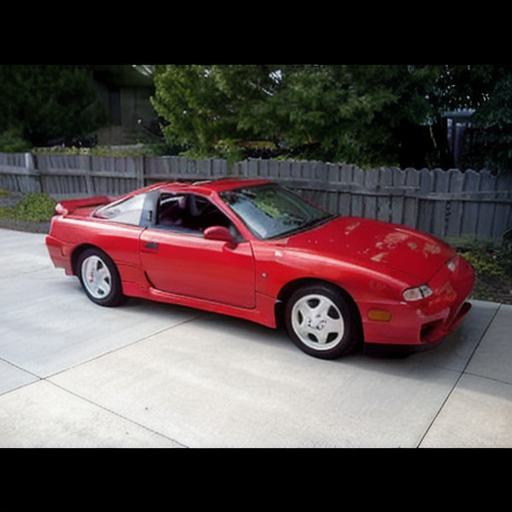} & \includegraphics[width=0.1\textwidth]{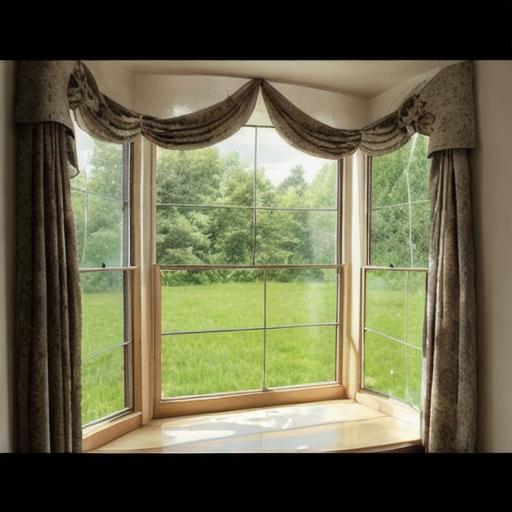} & \includegraphics[width=0.1\textwidth]{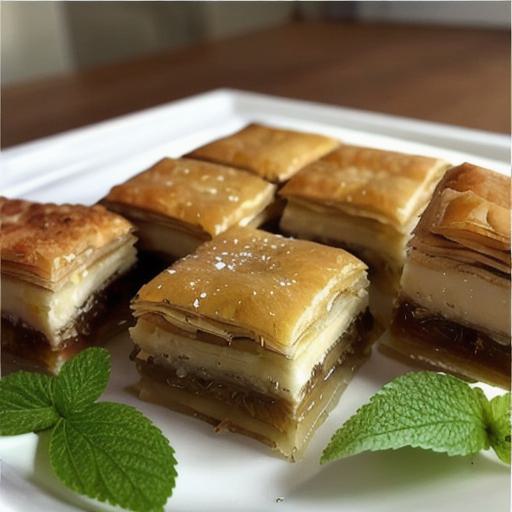} & \includegraphics[width=0.1\textwidth]{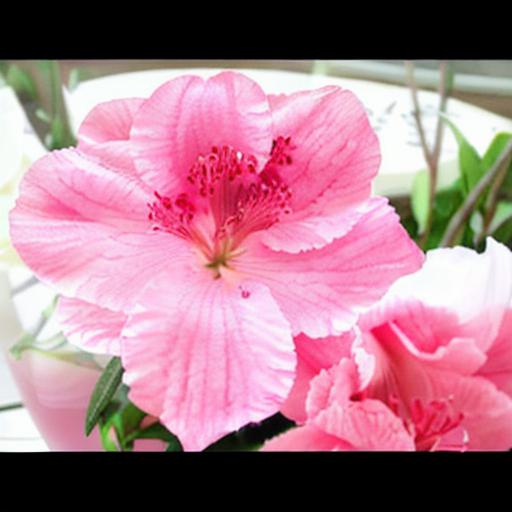} \\
\includegraphics[width=0.1\textwidth]{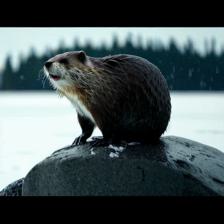} & \includegraphics[width=0.1\textwidth]{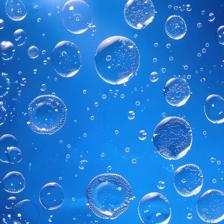} & \includegraphics[width=0.1\textwidth]{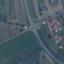} & \includegraphics[width=0.1\textwidth]{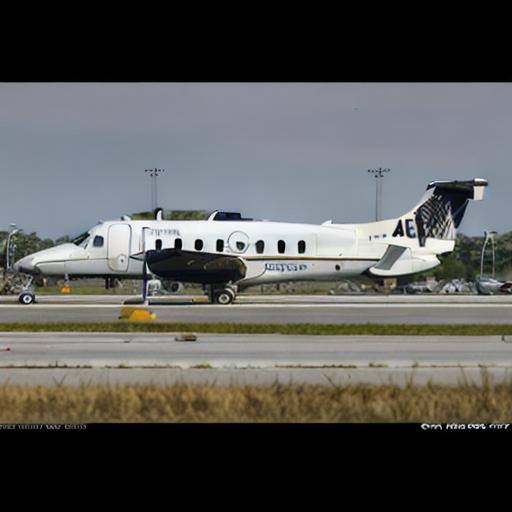} & \includegraphics[width=0.1\textwidth]{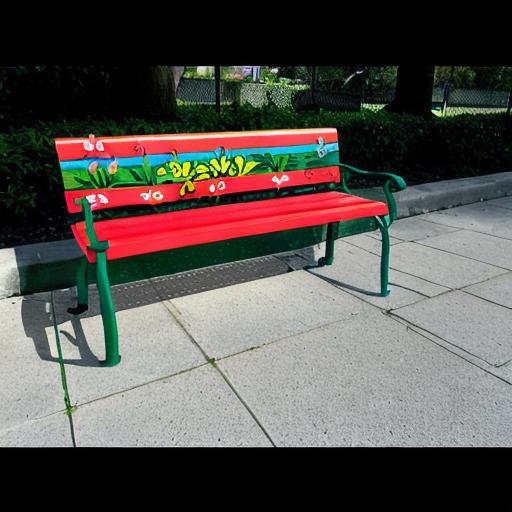} & \includegraphics[width=0.1\textwidth]{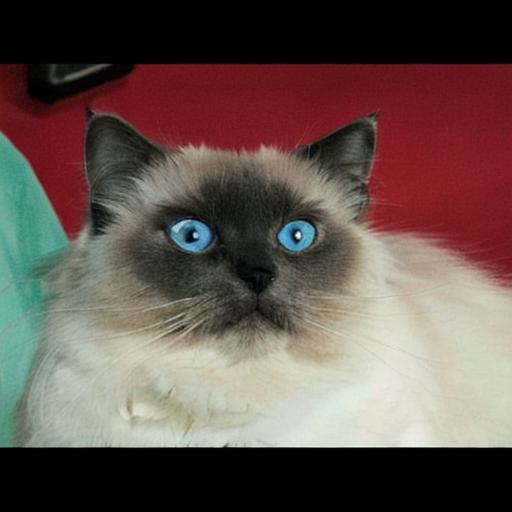} & \includegraphics[width=0.1\textwidth]{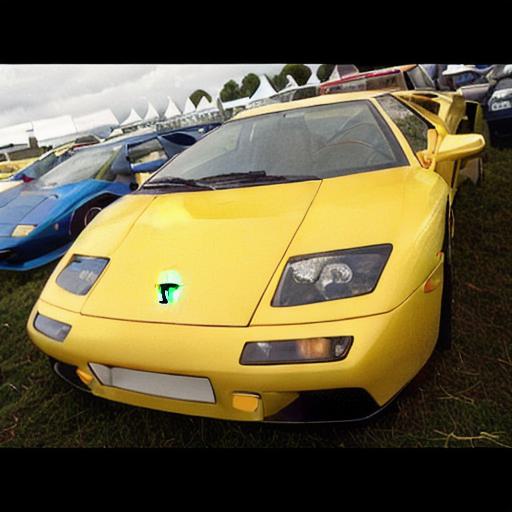} & \includegraphics[width=0.1\textwidth]{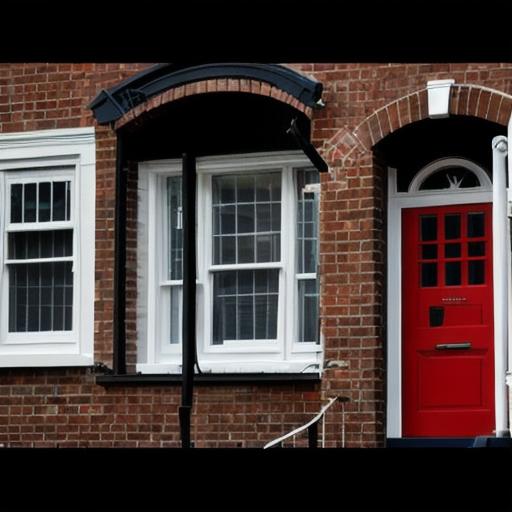} & \includegraphics[width=0.1\textwidth]{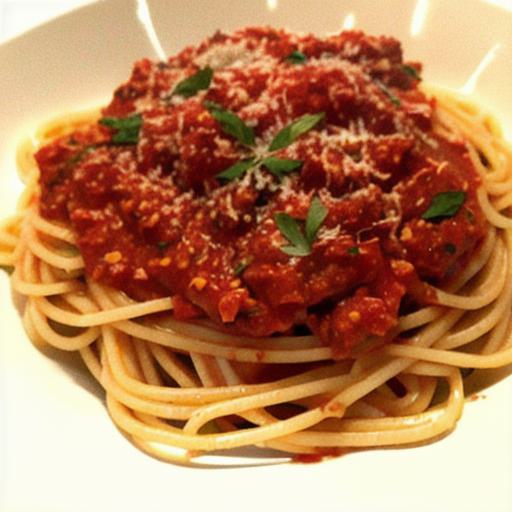} & \includegraphics[width=0.1\textwidth]{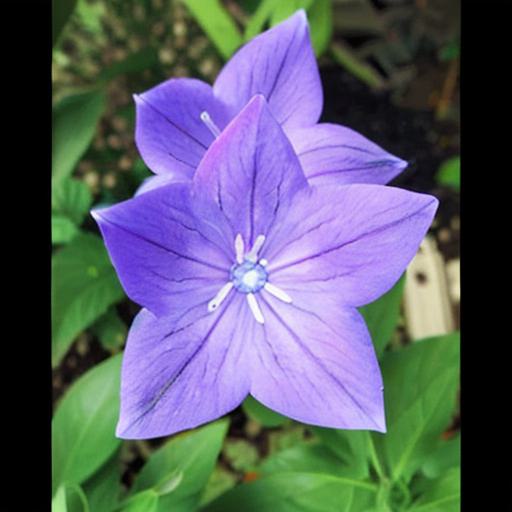} \\
\includegraphics[width=0.1\textwidth]{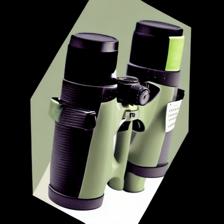} & \includegraphics[width=0.1\textwidth]{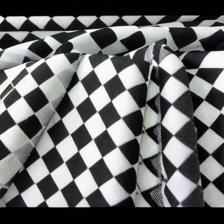} & \includegraphics[width=0.1\textwidth]{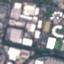} & \includegraphics[width=0.1\textwidth]{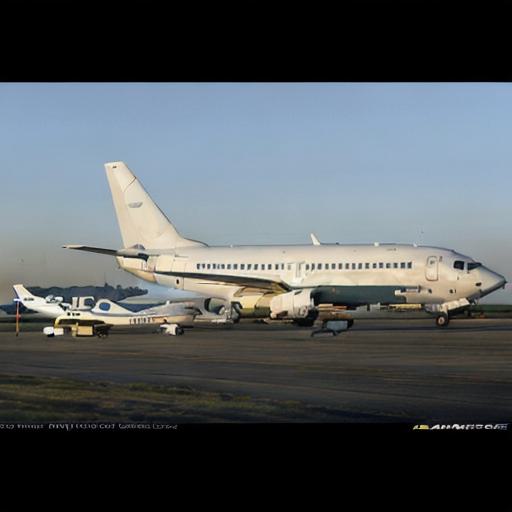} & \includegraphics[width=0.1\textwidth]{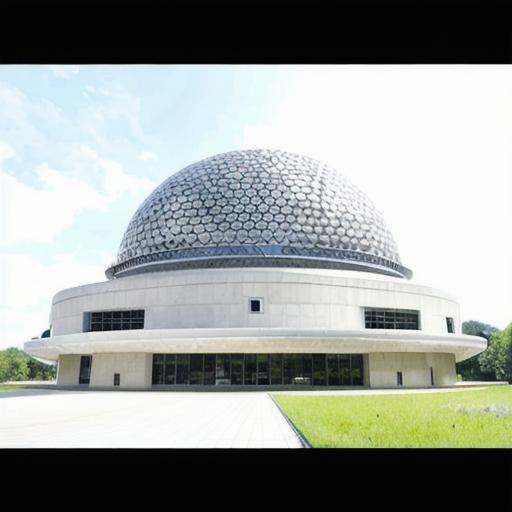} & \includegraphics[width=0.1\textwidth]{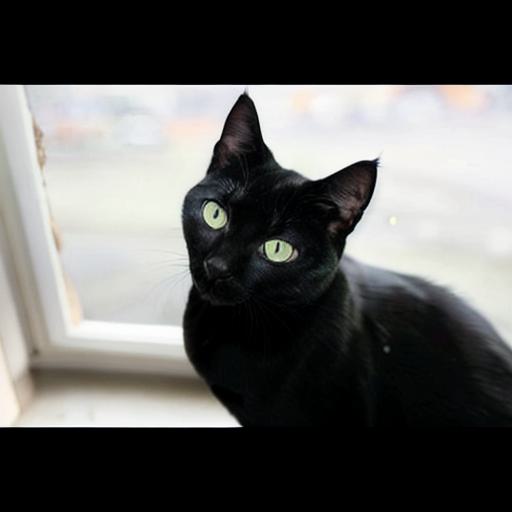} & \includegraphics[width=0.1\textwidth]{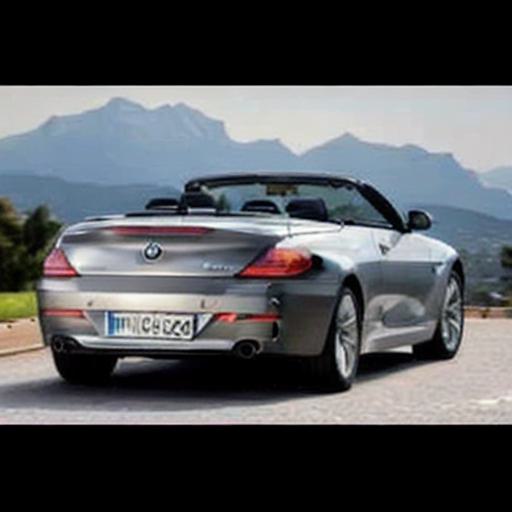} & \includegraphics[width=0.1\textwidth]{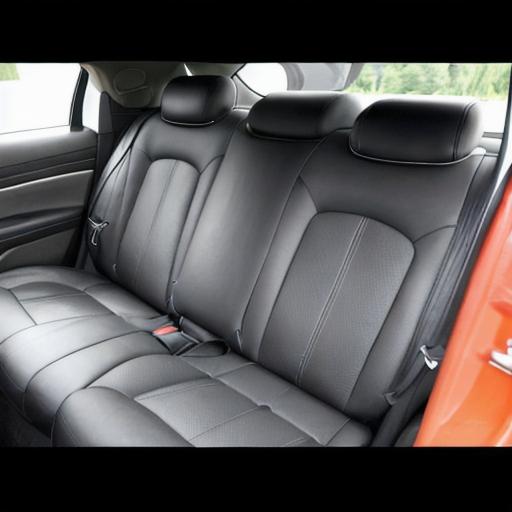} & \includegraphics[width=0.1\textwidth]{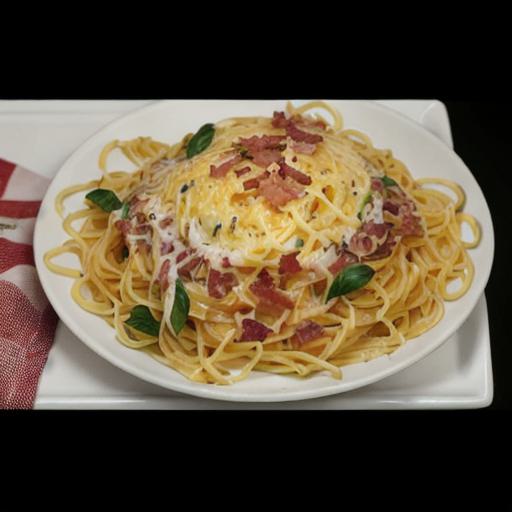} & \includegraphics[width=0.1\textwidth]{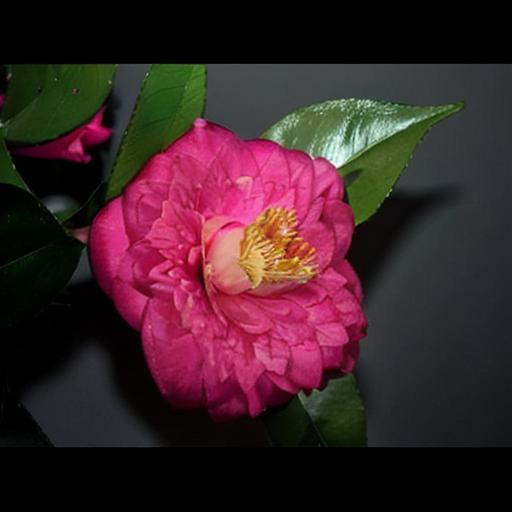} \\
\includegraphics[width=0.1\textwidth]{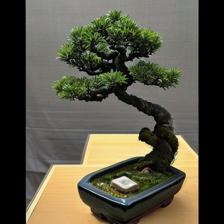} & \includegraphics[width=0.1\textwidth]{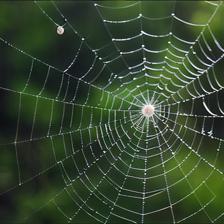} & \includegraphics[width=0.1\textwidth]{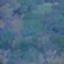}  & \includegraphics[width=0.1\textwidth]{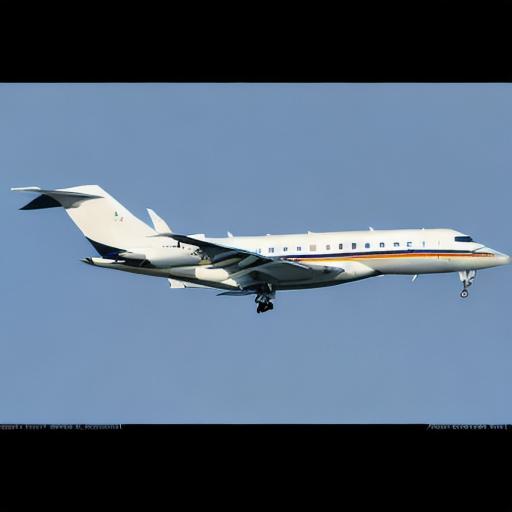} & \includegraphics[width=0.1\textwidth]{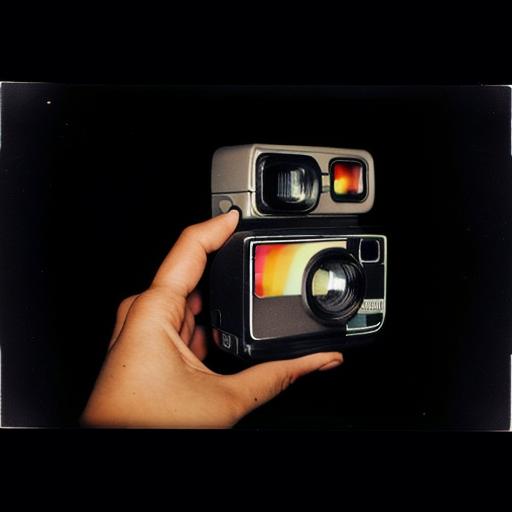} & \includegraphics[width=0.1\textwidth]{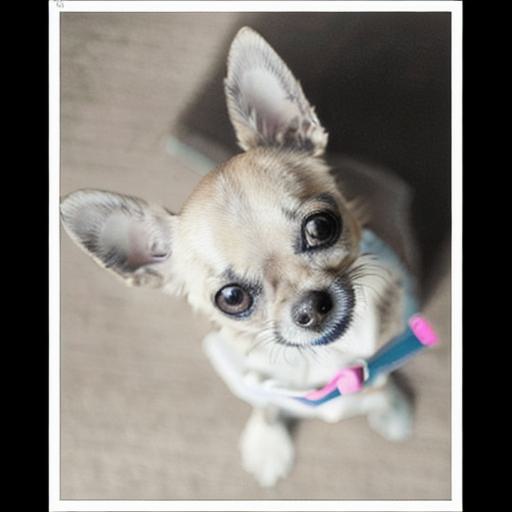} & \includegraphics[width=0.1\textwidth]{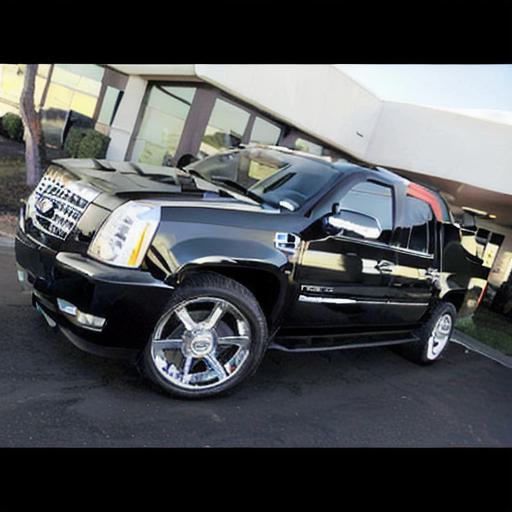} & \includegraphics[width=0.1\textwidth]{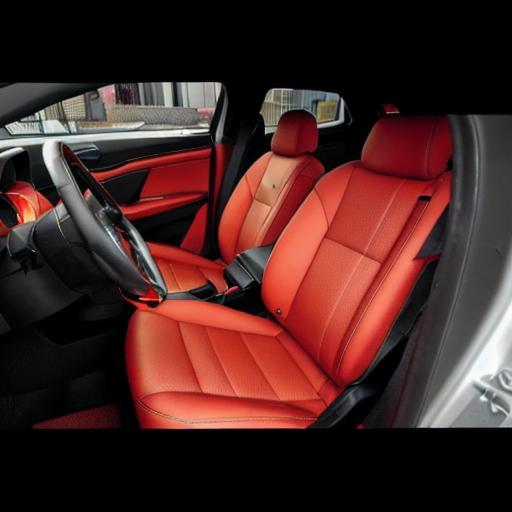} & \includegraphics[width=0.1\textwidth]{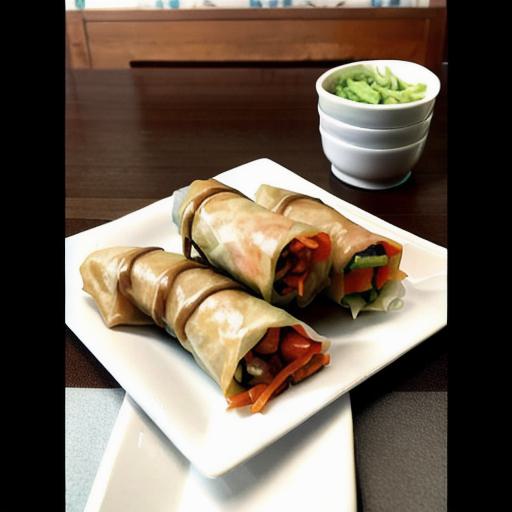} & \includegraphics[width=0.1\textwidth]{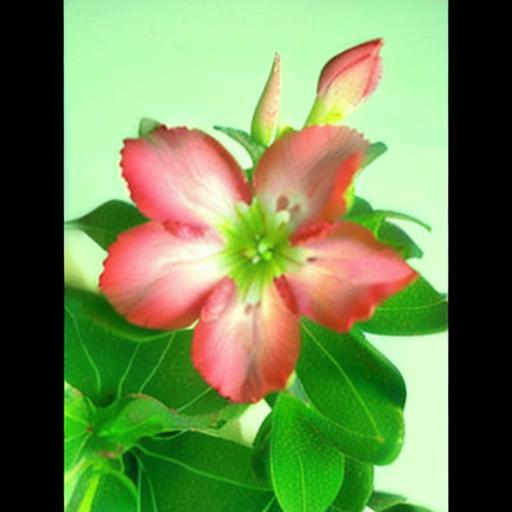} \\
\includegraphics[width=0.1\textwidth]{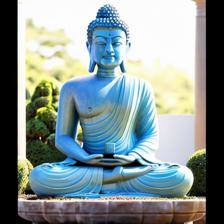} & \includegraphics[width=0.1\textwidth]{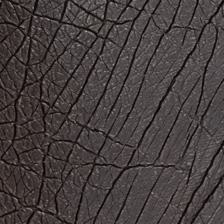} & \includegraphics[width=0.1\textwidth]{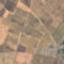}  & \includegraphics[width=0.1\textwidth]{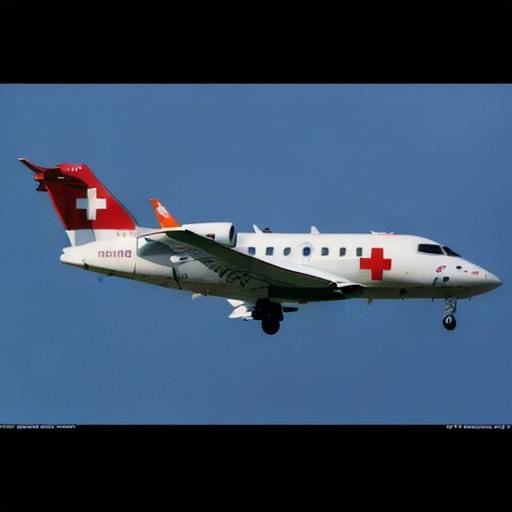} & \includegraphics[width=0.1\textwidth]{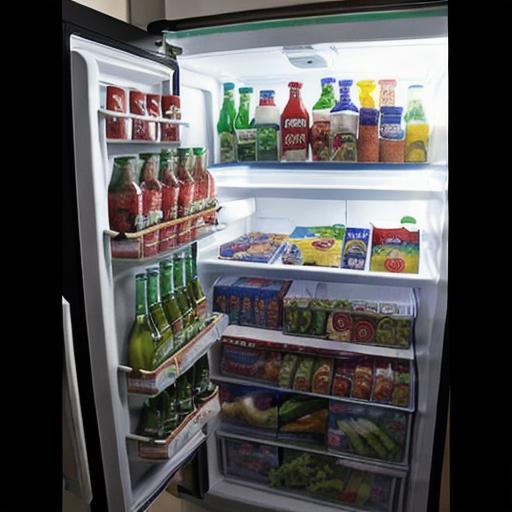} & \includegraphics[width=0.1\textwidth]{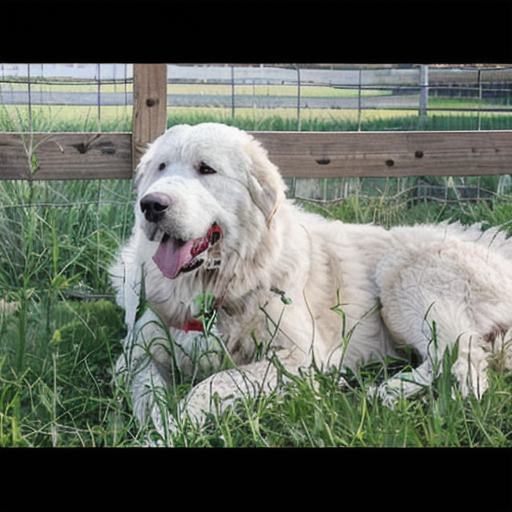} & \includegraphics[width=0.1\textwidth]{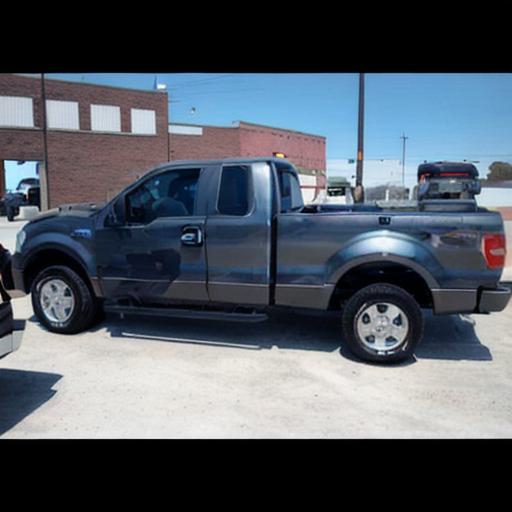} & \includegraphics[width=0.1\textwidth]{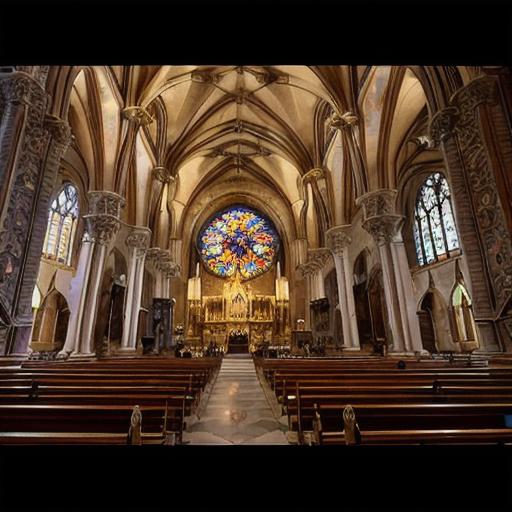} & \includegraphics[width=0.1\textwidth]{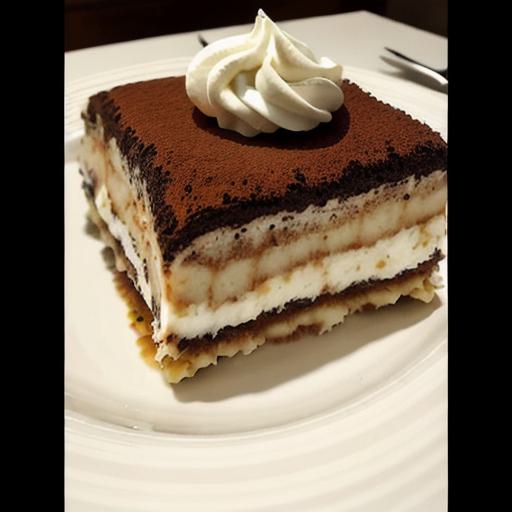} & \includegraphics[width=0.1\textwidth]{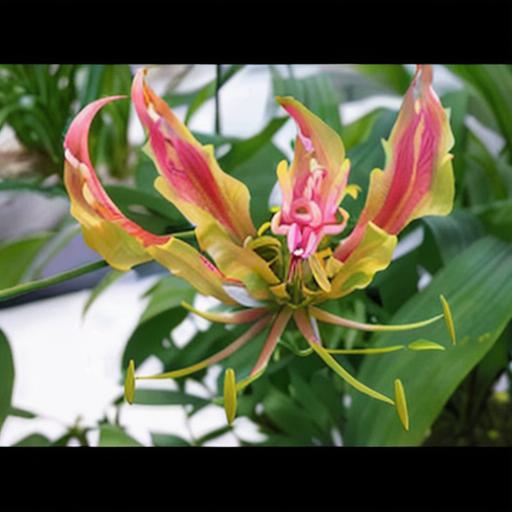} \\
\includegraphics[width=0.1\textwidth]{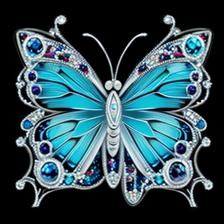} & \includegraphics[width=0.1\textwidth]{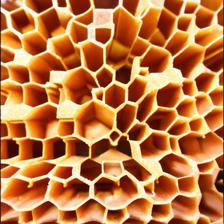} & \includegraphics[width=0.1\textwidth]{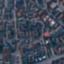}  & \includegraphics[width=0.1\textwidth]{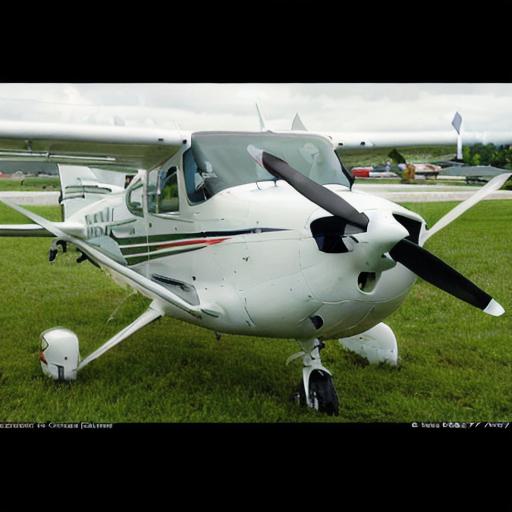} & \includegraphics[width=0.1\textwidth]{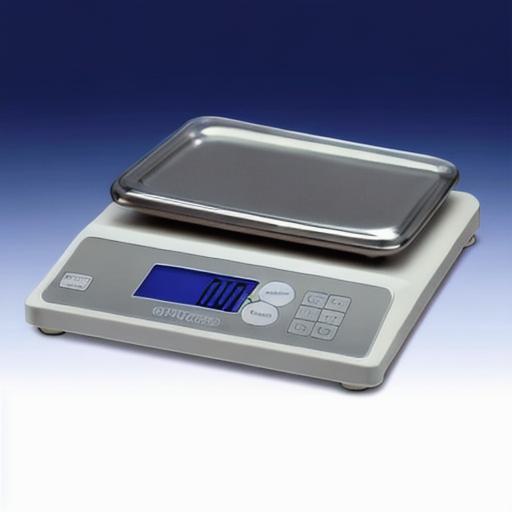} & \includegraphics[width=0.1\textwidth]{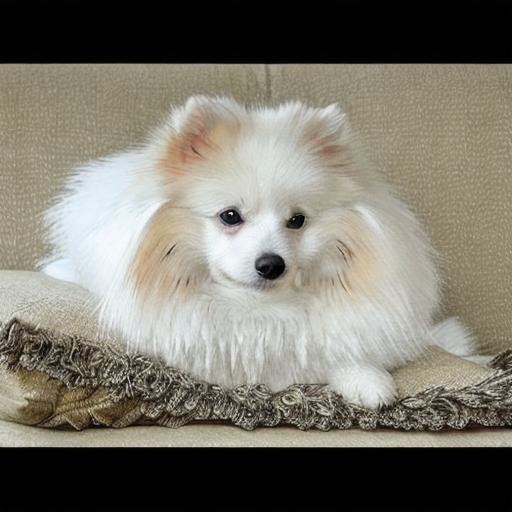} & \includegraphics[width=0.1\textwidth]{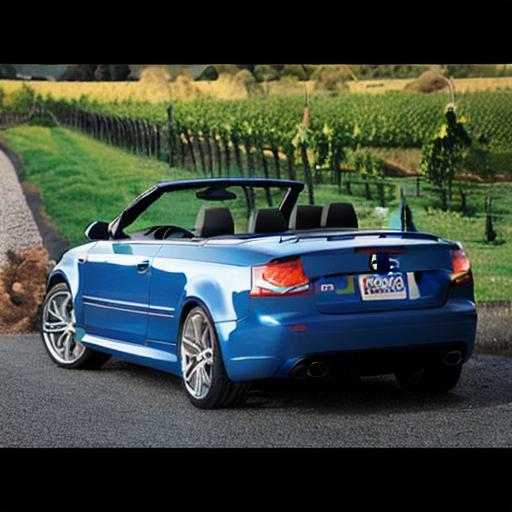} & \includegraphics[width=0.1\textwidth]{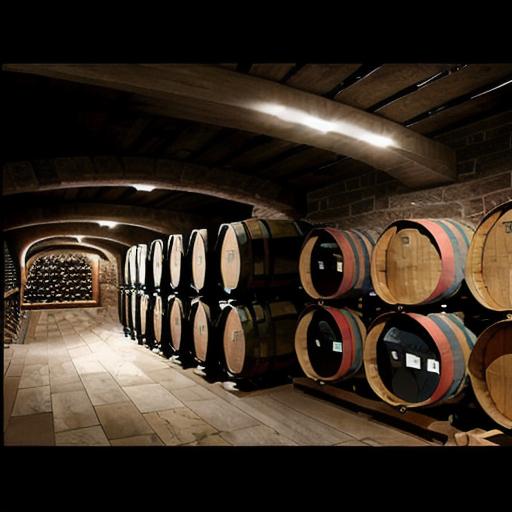} & \includegraphics[width=0.1\textwidth]{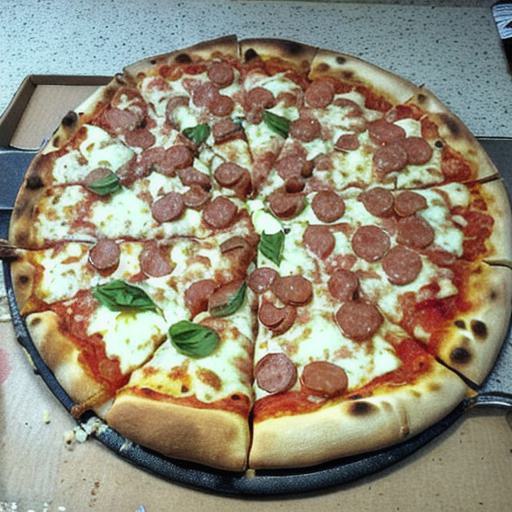}  & \includegraphics[width=0.1\textwidth]{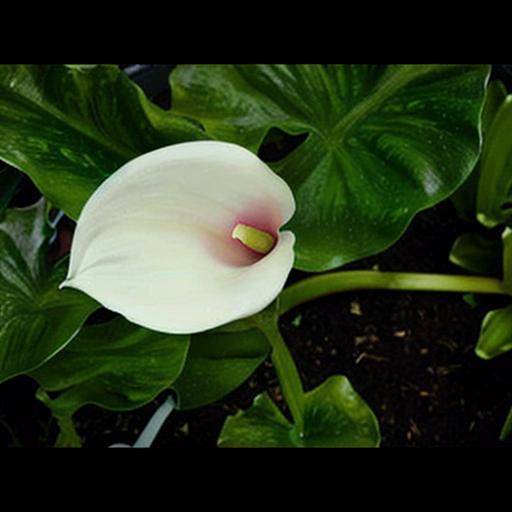} \\
\includegraphics[width=0.1\textwidth]{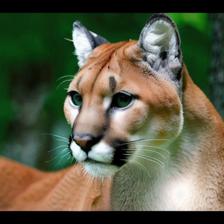} & \includegraphics[width=0.1\textwidth]{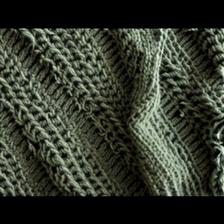} & \includegraphics[width=0.1\textwidth]{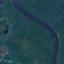}  & \includegraphics[width=0.1\textwidth]{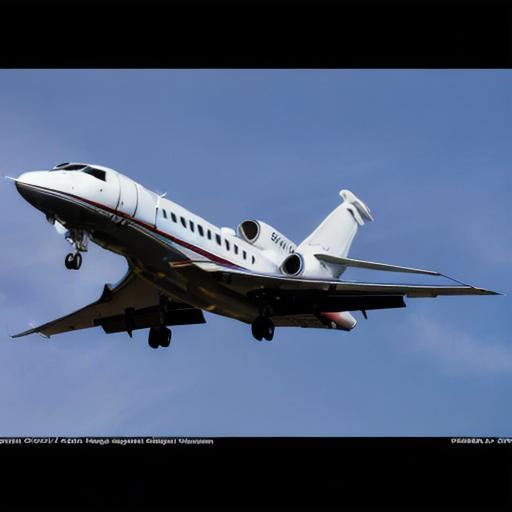} & \includegraphics[width=0.1\textwidth]{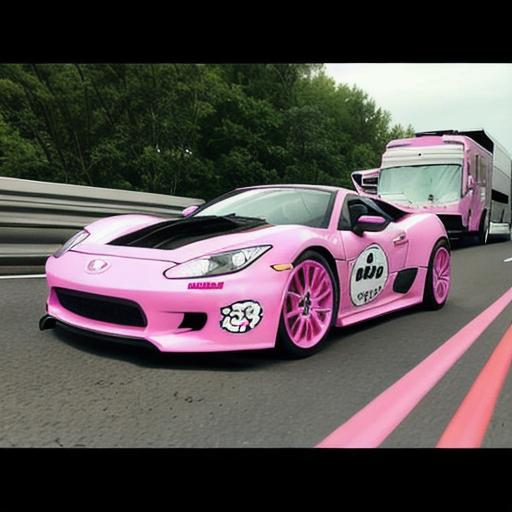} & \includegraphics[width=0.1\textwidth]{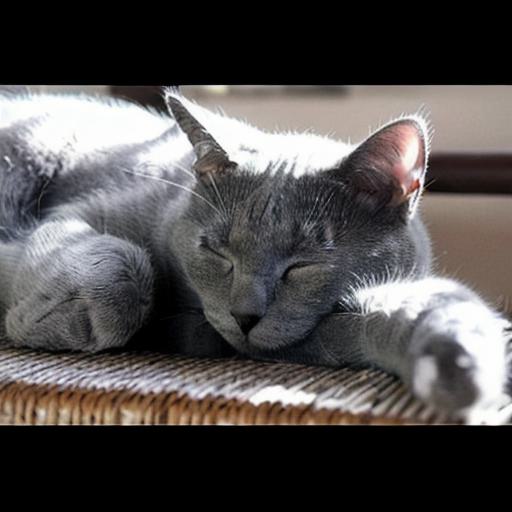} & \includegraphics[width=0.1\textwidth]{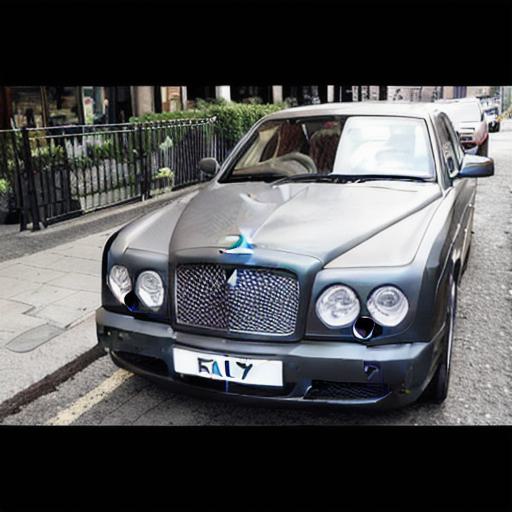} & \includegraphics[width=0.1\textwidth]{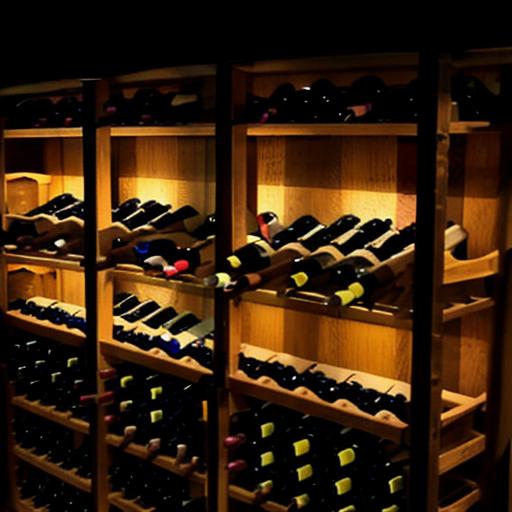} & \includegraphics[width=0.1\textwidth]{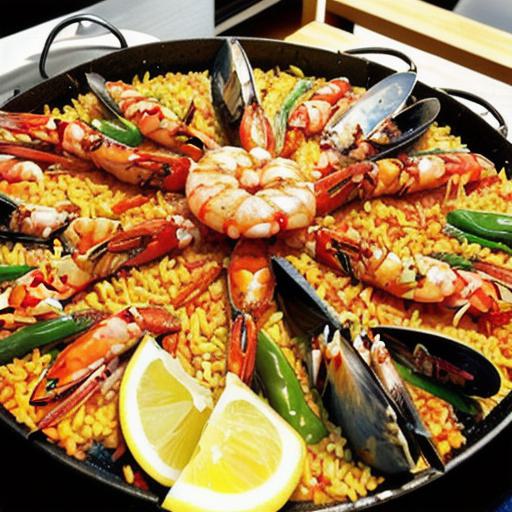} & \includegraphics[width=0.1\textwidth]{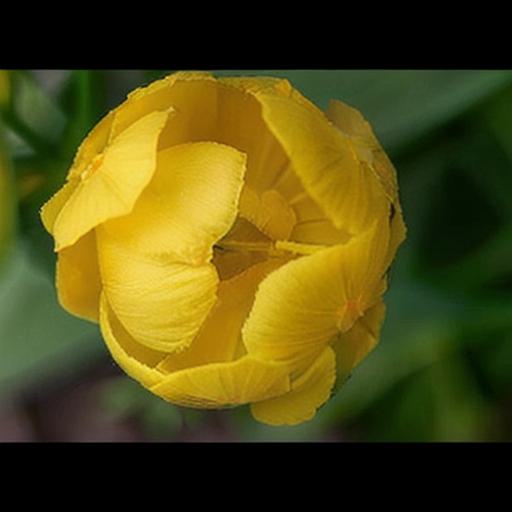} \\
\includegraphics[width=0.1\textwidth]{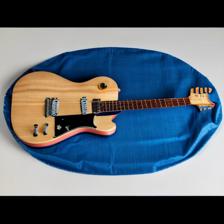} & \includegraphics[width=0.1\textwidth]{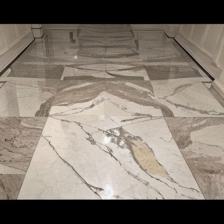} & \includegraphics[width=0.1\textwidth]{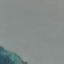}  & \includegraphics[width=0.1\textwidth]{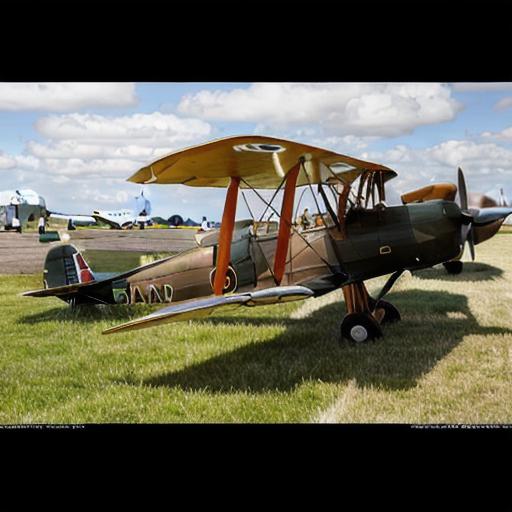} & \includegraphics[width=0.1\textwidth]{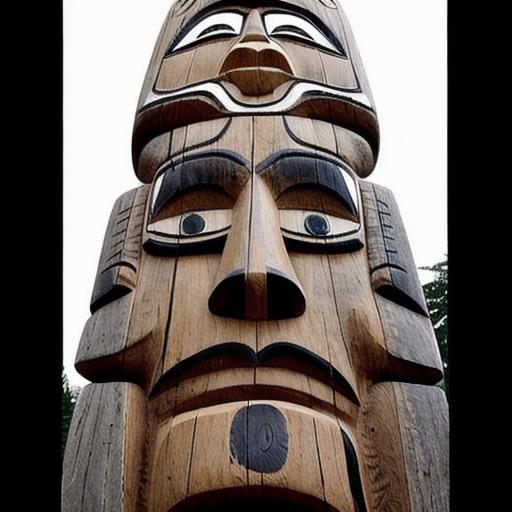} & \includegraphics[width=0.1\textwidth]{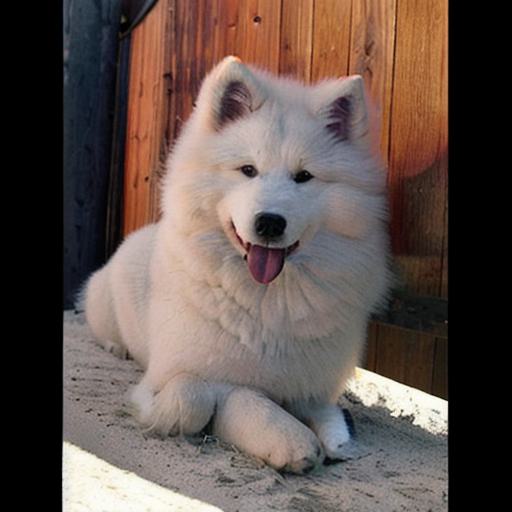} & \includegraphics[width=0.1\textwidth]{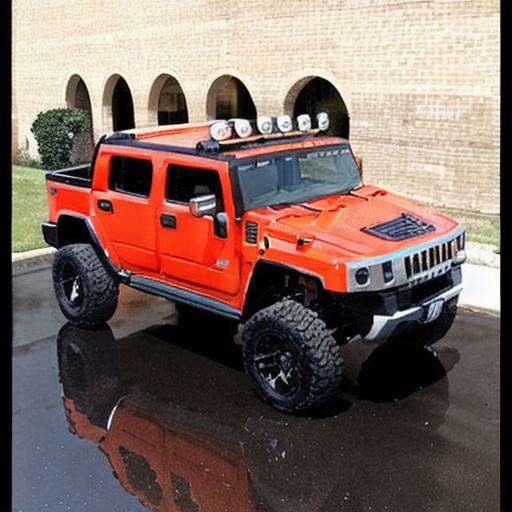} & \includegraphics[width=0.1\textwidth]{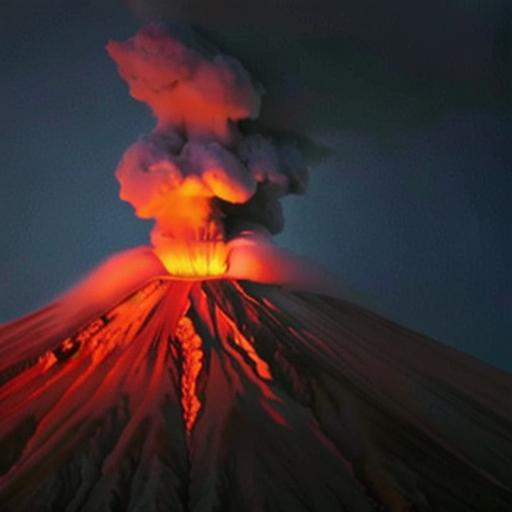} & \includegraphics[width=0.1\textwidth]{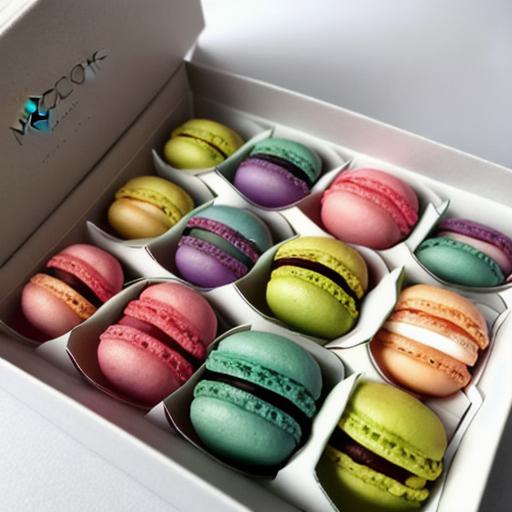} & \includegraphics[width=0.1\textwidth]{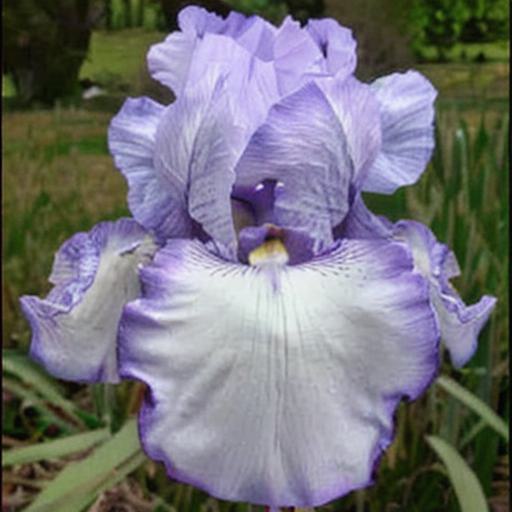} \\
\includegraphics[width=0.1\textwidth]{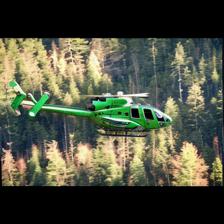} & \includegraphics[width=0.1\textwidth]{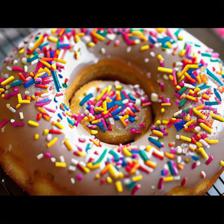} & \includegraphics[width=0.1\textwidth]{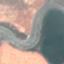}  & \includegraphics[width=0.1\textwidth]{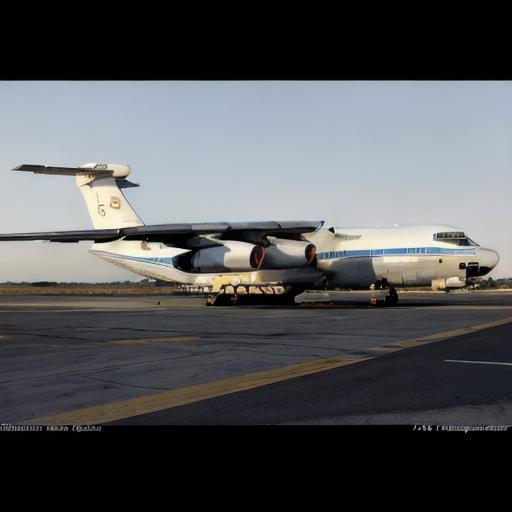} & \includegraphics[width=0.1\textwidth]{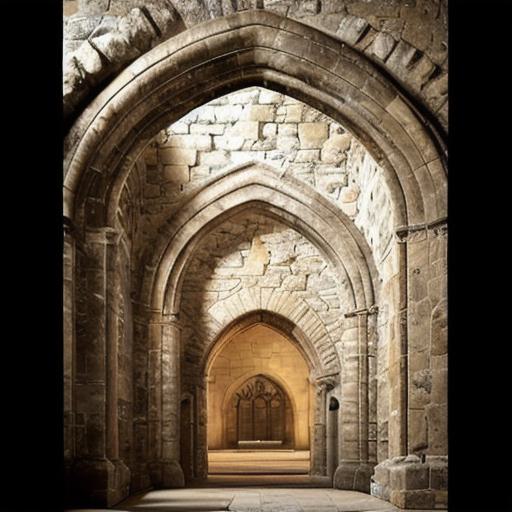} & \includegraphics[width=0.1\textwidth]{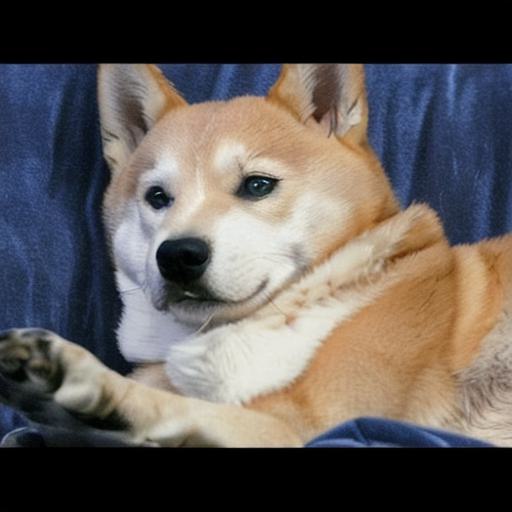} & \includegraphics[width=0.1\textwidth]{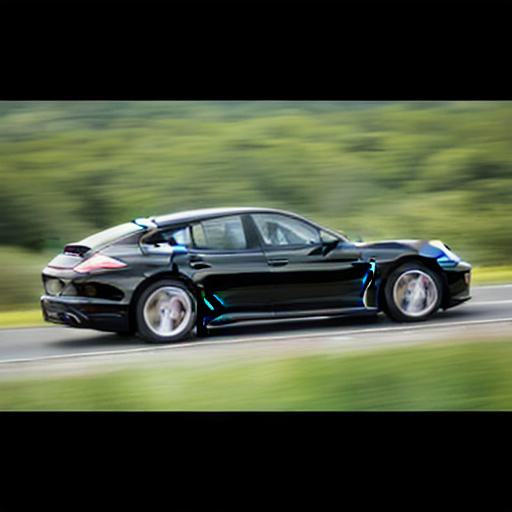} & \includegraphics[width=0.1\textwidth]{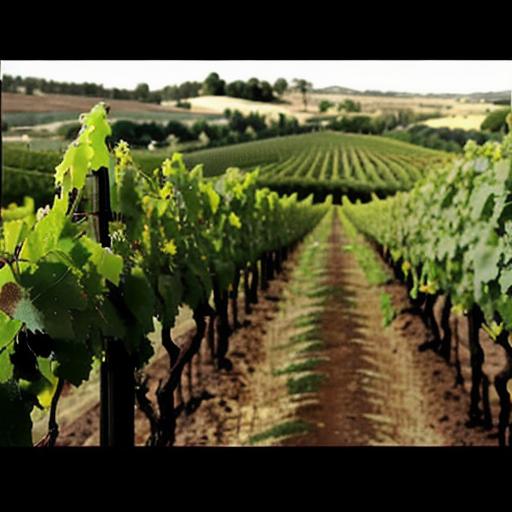} & \includegraphics[width=0.1\textwidth]{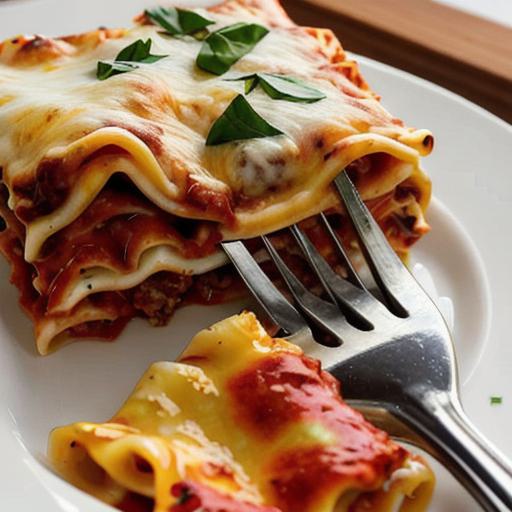} & \includegraphics[width=0.1\textwidth]{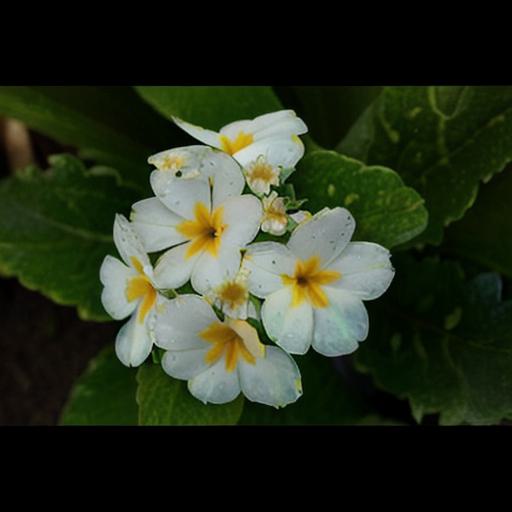} \\
\includegraphics[width=0.1\textwidth]{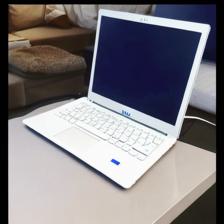} & \includegraphics[width=0.1\textwidth]{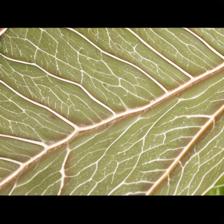} & \includegraphics[width=0.1\textwidth]{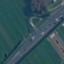}  & \includegraphics[width=0.1\textwidth]{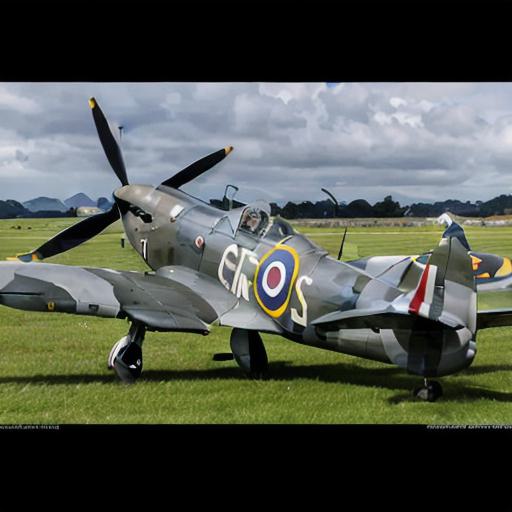} & \includegraphics[width=0.1\textwidth]{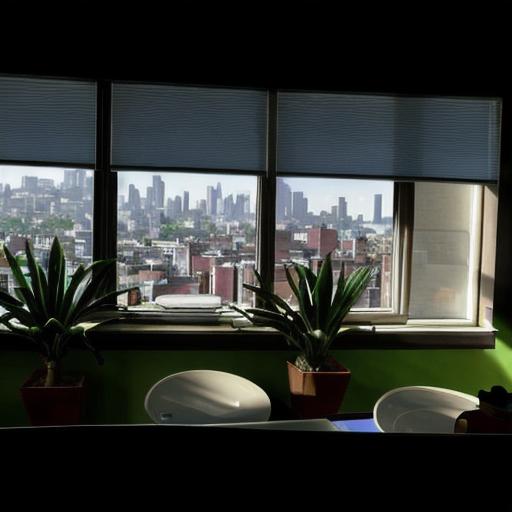} & \includegraphics[width=0.1\textwidth]{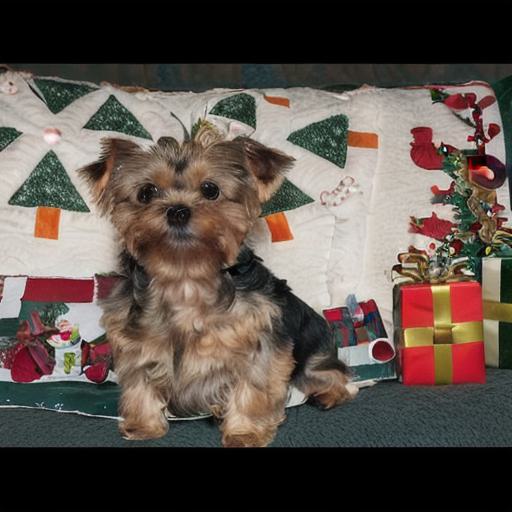} & \includegraphics[width=0.1\textwidth]{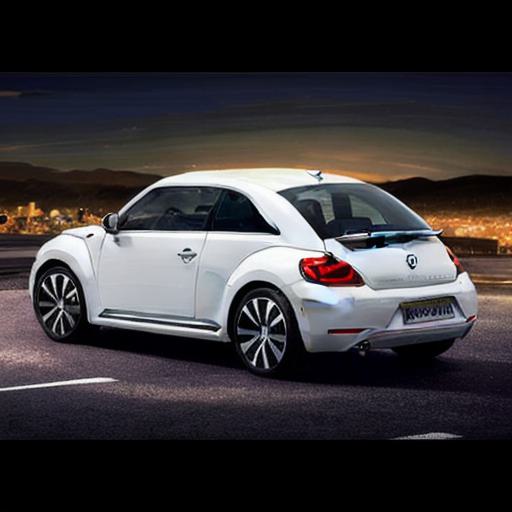} & \includegraphics[width=0.1\textwidth]{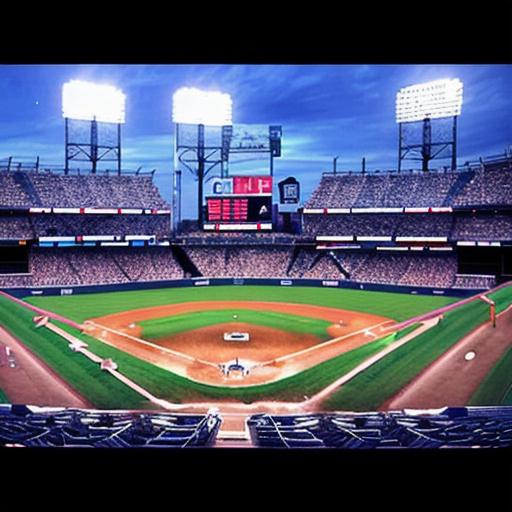} & \includegraphics[width=0.1\textwidth]{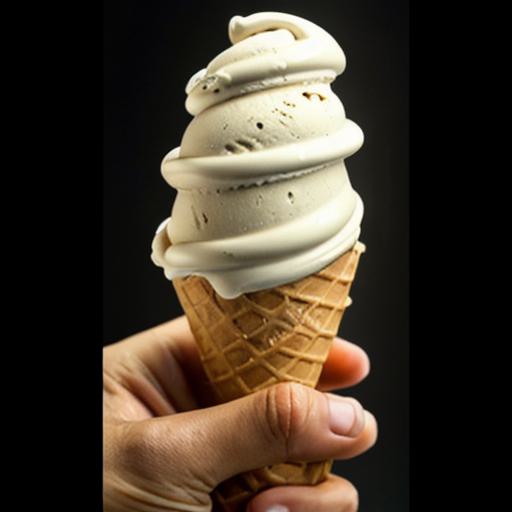} &  \includegraphics[width=0.1\textwidth]{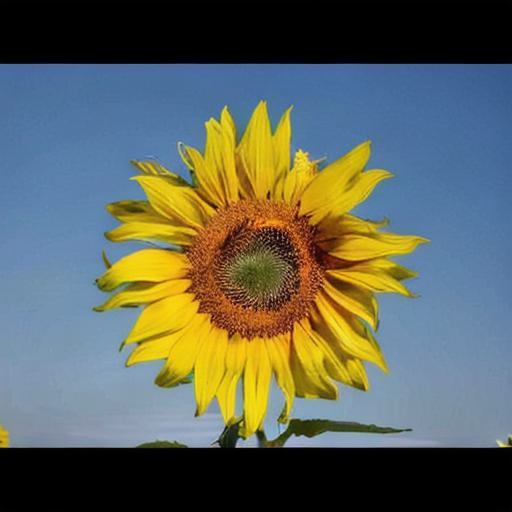}
\end{tabular}
\end{center}
\end{figure*}

We show more images generated by our SAP in Figure~\ref{fig:another_one} and Figure~\ref{fig:more_qualitatives}. Figure~\ref{fig:another_one} shows the in-domain synthesis capabilities of our SAP, while Figure~\ref{fig:more_qualitatives} shows randomly-picked samples for all the datasets, from left to right, Caltech101, DTD, Eurosat, FGVC Aircraft, ImageNet, Oxford Pets, Stanford Cars, SUN397, Food 101, and Flowers 102. 

In Figure~\ref{fig:another_one}, in the first row (highlighted with a red box), we should a real image from the 16 available shots for that class. Below them, we show images of the same class generated by our SAP. We can notice two different behaviors depending on the amount of noise injected. For the datasets where classes are more well-separated, \eg DTD and SUN397, we inject a higher amount of noise, as reported in Table~\ref{tab:parameters}. This leads to more diverse samples \wrt the original image, while the generated images maintain their semantic properties. For datasets where the classes are more overlapping in the visual domain, \eg EuroSAT or FGVC Aircraft, we prefer fidelity over diversity, as injecting too much noise might destroy semantic-relevant information, therefore for these datasets we choose to stop at step $5$ of the diffusion process. This leads to images which closely resemble original samples with variations in the details, \eg the livery of the plane.


In Figure~\ref{fig:more_qualitatives}, we show different classes (top to bottom) for each dataset (left to right), providing the corresponding class names in Table~\ref{tab:qualitatives_classnames}. Overall, the generated images exhibit high fidelity and a high degree of realism, even for the dataset where we inject a high degree of noise, \eg Caltech101, DTD, ImageNet, SUN397, and Food 101 (we refer to Table~\ref{tab:parameters} for more details).

A detailed inspection of the images highlights some hallucinations in the generation, \eg the shape of the helicopter in the eleventh row of Caltech101, the missing hole in the donut in the eleventh row of DTD, the hallucinated small plane in the fifth row of FGVC Aircraft, or the weird hand posture in the sixth row of ImageNet. Although these would constitute errors in traditional generative tasks, for the recognition tasks, we deem these inconsistencies as not influential as the semantics of the original class are preserved. Nevertheless, our SAP can fully exploit the rich LLaVA captions to generate high-quality, realistic, and detail-rich images of the desired class. Moreover, the use of real samples as the starting point for the generation allows the model to better preserve the semantics of each class, \eg the proportion of the planes in FGVC Aircraft. Additionally, our generation is able to distinguish and generate very similar classes, \eg it can distinguish between bolognese spaghetti and carbonara spaghetti (fourth and fifth row of Food 101, although one could argue basil does not belong on carbonara).

\begin{table*}[!ht]
\caption{Class names of the generated images in Figure~\ref{fig:more_qualitatives}.}
\label{tab:qualitatives_classnames}
\resizebox{1.\textwidth}{!}{
\begin{tabular}{@{}cccccccccc@{}}
\toprule
Caltech         & DTD         & EuroSAT               & FGVC Aircraft                       & ImageNet     & Oxford Pets       & Stanford Cars           & SUN397                 & Food 101       & Flowers 102           \\ \midrule
Anchor          & Banded      & AnnualCrop            & Airbus A300                         & Tench        & Abyssinian        & 1991 Volkswagen Golf    & Balcony Exterior       & Apple Pie      & Japanese Anemone      \\
Ant             & Blotchy     & Forest                & Antonov An-12                       & Brambling    & American Bulldog  & 1993 Geo Metro          & Balcony Interior       & Baby Back Ribs & Thorn Apple           \\
Barrel          & Braided     & Herbaceous Vegetation & ATR-72                              & Palace       & American Pitbull  & 1998 Nissan 240SX       & Bow Window Indoor      & Baklava        & Azalea                \\
Beaver          & Bubbly      & Highway               & Beechcraft 1900                     & Bench        & Birman            & 2001 Lamborghini Diablo & Bow Window Outdoor     & Bolognese      & Balloon Flower        \\
Binoculars      & Chequered   & Industrial            & Boeing 737-200                      & Planetarium  & Bombay            & 2007 BMW Serie 6        & Car Interior Backseat  & Carbonara      & Camelia               \\
Bonsai          & Cobwebbed   & Pasture               & Bombardier Aerospace Global Express & Camera       & Chihuahua         & 2007 Cadillac Escalade  & Car Interior Frontseat & Spring Rolls   & Desert Rose           \\
Buddha          & Cracked     & Permanent Crop        & Canadair Challenger 600             & Fridge       & Great Pyrenees    & 2007 Ford F-150         & Cathedral Indoor       & Tiramisu       & Fire Lily             \\
Butterfly       & Honeycombed & Residential           & Cessna 172                          & Scale        & Pomeranian        & 2008 Audi RS4           & Wine Barrel Storage    & Pizza          & Giant White Arum Lily \\
Cougar          & Interlaced  & River                 & Dassault Aviation Falcon 900        & Sport Car    & Russian Blue      & 2009 Bentley Arnage     & Wine Bottle Storage    & Paella         & Globe Flower          \\
Electric Guitar & Marbled     & SeaLake               & de Havilland DH-82                  & Totem Pole   & Samoyed           & 2009 Hummer H2          & Volcano                & Macarons       & Bearded Iris          \\
Helicopter      & Sprinkled   & River                 & Ilyushin Il-76                      & Vault        & Shiba Inu         & 2012 Porsche Panamera   & Vineyard               & Lasagna        & Primula               \\
Laptop          & Veined      & Highway               & Supermarine Spitfire                & Window Shade & Yorkshire Terrier & 2012 Volkswagen Beetle  & Baseball Field         & Ice Cream      & Sunflower             \\ \bottomrule
\end{tabular}
}
\end{table*}
\begin{table*}[!htb]
\caption{Per-dataset parameters of \methodname}
\resizebox{0.99\textwidth}{!}{
\begin{tabular}{@{}l|cccccccccc@{}}
\toprule
                  & Caltech101 & DTD & EuroSAT & FGVC Aircraft & ImageNet & Oxford Pets & Stanford Cars & SUN397 & Food 101 & Flowers 102 \\ \midrule

Model &  ViT-B/16 & ViT-B/16 & ViT-B/16 & ViT-B/16 & ViT-B/16 & ViT-B/16 & ViT-B/16 & ViT-B/16 & ViT-B/16 & ViT-B/16  \\
Optimizer & AdamW & AdamW & AdamW & AdamW & AdamW & AdamW & AdamW & AdamW & AdamW & AdamW  \\
Batch size $bs$    &    128      &  16   &     16    & 16     &    256      &  128    &  16   &  128   &   64   &    32     \\
Learning rate $lr$ &     $2^{-12}$     & $2^{-15}$    &     $2^{-12}$    &   $2^{-11}$   &    $2^{-15}$     &   $2^{-15}$   &  $2^{-13}$    &  $2^{-14}$   &   $2^{-14}$   &    $2^{-12}$     \\
Weight decay &  0.001 & 0.001 & 0.001 & 0.001 & 0.001 & 0.001 & 0.001 & 0.001 & 0.001 & 0.001  \\
$lr$ scheduler &  Cosine & Cosine & Cosine & Cosine & Cosine & Cosine & Cosine & Cosine & Cosine & Cosine  \\
Real/synth weight $\lambda$ &  0.8/0.2       &  0.5/0.5   & 0.8/0.2        &   0.8/0.2   &     0.8/0.2     &  0.7/0.3    &   0.8/0.2   &   0.8/0.2  &   0.5/0.5   &   0.8/0.2      \\
Vision Encoder LoRA $r$ &          64    &  64   &   16      &  64    &     64     & 16    &   64   &  64   &    64 &    64     \\
Vision Encoder LoRA $\alpha$ &         64    &   64  &   32      &   32   &     64     &   32   &   64   &   32  & 32     &     32    \\
Vision Encoder LoRA dropout & 0.1 & 0.1 & 0.1 & 0.1 & 0.1 & 0.1 & 0.1 & 0.1 & 0.1 & 0.1  \\
Text Encoder LoRA $r$ & 16 & 16 & 16 & 16 & 16 & 16 & 16 & 16 & 16 & 16  \\
Text Encoder LoRA $\alpha$ & 32 & 32 & 32 & 32 & 32 & 32 & 32 & 32 & 32 & 32\\
Text Encoder LoRA dropout & 0.1 & 0.1 & 0.1 & 0.1 & 0.1 & 0.1 & 0.1 & 0.1 & 0.1 & 0.1  \\

Augmentation    &      True     &  True   &   True   &  True    &    True      & True & True     &  True   &   True   &    True     \\
Cutmix    &     0.1     &   0.1  &   0.1    &   0.1   &     0.1     &  0.0   &   0.1   &  0.1   &  0.1   &    0.1      \\
Mixup    &     0.1     &   0.1  &   0.1    &   0.1   &     0.1     &  0.0   &   0.1   &  0.1   &  0.1   &    0.1      \\
Label-smoothing  &    0.1      &   0.1  &    0.1     &   0.1   &     0.1     &   0.0   &   0.1   &  0.1   &   0.1   &     0.1    \\
Diffusion Sampler & DPM-Solver++ & DPM-Solver++ & DPM-Solver++ & DPM-Solver++ & DPM-Solver++ & DPM-Solver++ & DPM-Solver++ & DPM-Solver++ & DPM-Solver++ & DPM-Solver++ \\
CFG Strength & 8 & 8 & 8 & 8 & 8 & 8 & 8 & 8 & 8 & 8 \\ 
Number of diffusion steps &  20 & 20 & 20 & 20 & 20 & 20 & 20 & 20 & 20 & 20 \\
Number of noising steps    &   5       &  5   &   15      &  15    &     5     &   15   &    15  &  5   &    15  &   15      \\ 
\bottomrule
\end{tabular}
}
\label{tab:parameters}
\end{table*}


\end{document}